\documentclass[lettersize,journal]{IEEEtran}
\usepackage{amsmath,amsfonts}
\usepackage{algorithmic}
\usepackage{array}
\usepackage[caption=false,font=normalsize,labelfont=sf,textfont=sf]{subfig}
\usepackage{textcomp}
\usepackage{stfloats}
\usepackage{url}
\usepackage{verbatim}
\usepackage{graphicx}
\usepackage{multirow}
\usepackage{booktabs}
\usepackage{enumitem}
\usepackage{amsmath} 
\usepackage[colorlinks=true, linkcolor=blue, citecolor=blue, urlcolor=blue]{hyperref}
\usepackage{graphicx} 
\usepackage{float}    
\usepackage{tabularx} 
\usepackage{amssymb} 
\usepackage{pifont}  
\usepackage{ragged2e} 
\usepackage{bm}
\usepackage{booktabs}
\usepackage{caption}
\newcommand{\cmark}{\checkmark}
\newcommand{\xmark}{\ding{55}}
\def\BibTeX{{\rm B\kern-.05em{\sc i\kern-.025em b}\kern-.08em
    T\kern-.1667em\lower.7ex\hbox{E}\kern-.125emX}}
\usepackage{balance}

\begin{document}
\title{Spectral-Spatial Self-Supervised Learning for Few-Shot Hyperspectral Image Classification}

\author{
    \IEEEauthorblockN{ 
        Wenchen Chen\IEEEauthorrefmark{1}, 
        Yanmei Zhang\IEEEauthorrefmark{1},
        Zhongwei Xiao\IEEEauthorrefmark{1}, 
        Jianping Chu\IEEEauthorrefmark{1}, 
        and Xingbo Wang\IEEEauthorrefmark{1}
    }

    
    \thanks{\IEEEauthorrefmark{1}All authors are with Beijing Institute of Technology, Beijing 100081, China.} 
    
    \thanks{Wenchen Chen's e-mail: 3120221303@bit.edu.cn.}
    \thanks{Zhongwei Xiao's e-mail: xzw0321057@outlook.com.}
    \thanks{Jianping Chu's e-mail: 3120221308@bit.edu.cn.}
    \thanks{Xingbo Wang's e-mail: 3220221535@bit.edu.cn.}
    \thanks{Corresponding author: Yanmei Zhang (e-mail: 0726zym@bit.edu.cn).} 
}

\markboth{IEEE TRANSACTIONS ON GEOSCIENCE AND REMOTE SENSING, VOL. XX, NO. YY, MONTH YEAR}{CHEN \MakeLowercase{\textit{et al.}}: SPECTRAL-SPATIAL SELF-SUPERVISED LEARNING FOR HSI CLASSIFICATION}

\maketitle

\begin{abstract}
  Few-shot classification of hyperspectral images (HSI) faces the challenge of scarce labeled samples. Self-Supervised learning (SSL) and Few-Shot Learning (FSL) offer promising avenues to address this issue. However, existing methods often struggle to adapt to the spatial geometric diversity of HSIs and lack sufficient spectral prior knowledge. To tackle these challenges, we propose a method, Spectral-Spatial Self-Supervised Learning for Few-Shot Hyperspectral Image Classification (S4L-FSC), aimed at improving the performance of few-shot HSI classification.
Specifically, we first leverage heterogeneous datasets to pretrain a spatial feature extractor using a designed Rotation-Mirror Self-Supervised Learning (RM-SSL) method, combined with FSL. This approach enables the model to learn the spatial geometric diversity of HSIs using rotation and mirroring labels as supervisory signals, while acquiring transferable spatial meta-knowledge through few-shot learning. Subsequently, homogeneous datasets are utilized to pretrain a spectral feature extractor via a combination of FSL and Masked Reconstruction Self-Supervised Learning (MR-SSL). The model learns to reconstruct original spectral information from randomly masked spectral vectors, inferring spectral dependencies. In parallel, FSL guides the model to extract pixel-level discriminative features, thereby embedding rich spectral priors into the model.
This spectral-spatial pretraining method, along with the integration of knowledge from heterogeneous and homogeneous sources, significantly enhances model performance. Extensive experiments on four HSI datasets demonstrate the effectiveness and superiority of the proposed S4L-FSC approach for few-shot HSI classification. The code for this research is publicly available at https://github.com/Wenchen-Chen/S4L-FSC.
\end{abstract}

\begin{IEEEkeywords}
Few-shot learning (FSL), hyperspectral image (HSI) classification, self-supervised learning(SSL), contrastive learning(CL).
\end{IEEEkeywords}

\section{Introduction}
\IEEEPARstart{H}{yperspectral} image (HSI) classification is a core research area in remote sensing. It receives significant attention. This is due to its wide applications in land cover classification, resource exploration, and environmental monitoring \cite{goetz1985imaging,LIANG2015123,6555921}. However, annotating HSI data is time-consuming. It is also costly. This leads to a scarcity of training samples. This scarcity greatly restricts the use of deep learning models in HSI classification. This limitation is particularly evident in tasks that require high-precision modeling of spatio-spectral features \cite{kumar2020feature}.

In early HSI classification research, scholars primarily used traditional machine learning techniques. Examples include Support Vector Machines (SVM)~\cite{1323134}, K-Nearest Neighbors (K-NN) \cite{8515116}, and Random Forest \cite{ham2005investigation}. These methods performed well when many samples were available. However, they struggled in few-shot scenarios. These methods relied on handcrafted features and shallow models. Thus, they could not effectively capture the complex spatio-spectral characteristics of HSI. This resulted in a sharp decline in their generalization ability \cite{jia2021survey}.

As deep learning technology advanced, early applications in HSI classification mainly focused on extracting spectral features. Chen et al. proposed an HSI classification method using Stacked Auto-Encoders (SAE) \cite{6844831}. This method used unsupervised learning. It extracted deep features from high-dimensional spectral data. Classification was then performed using logistic regression. The SAE method achieved higher accuracy on public datasets like Indian Pines than traditional methods. It became a pioneering work for applying deep learning to HSI classification. Later, methods based on Deep Belief Networks (DBN) were also introduced \cite{7026039}. Tao et al. used DBNs for hierarchical modeling of spectral information. This further improved classification performance. These approaches formed the basis for deep learning in HSI classification. However, they primarily considered spectral information. They also ignored spatial context. This limited their performance in complex scenarios.

After the introduction of AlexNet \cite{krizhevsky2012imagenet} and ResNet \cite{he2016deep}, deep learning-based models were gradually applied to HSI classification. These models use end-to-end feature learning mechanisms. They can automatically extract joint spatio-spectral features from hyperspectral data. This significantly improved classification accuracy. For example, methods such as 3DCNN \cite{li2017spectral}, SSRN \cite{zhong2017spectral}, and HybridSN \cite{roy2019hybridsn} can achieve near-perfect classification. This is possible even when using only 20\% of the overall data as training samples. With the rise of the Transformer architecture, researchers began exploring its potential for HSI classification. For instance, Zhong et al. proposed a Spectral-Spatial Transformer Network (SSTN) \cite{zhong2021spectral}. SSTN combined spatial attention with spectral correlation modules. This design overcame the geometric limitations of traditional convolutional kernels. It also enabled long-range feature interaction modeling.

Despite these advancements, the dependence of deep learning models on large amounts of labeled data remains a bottleneck. This hinders their widespread application in HSI classification. Labeled samples in HSI datasets are typically limited. Consequently, models are prone to overfitting. This makes it difficult for them to achieve their full performance potential in practical scenarios \cite{paoletti2019deep,li2018data,Prototypical_net}.

To address the problem of insufficient labeled samples, researchers have explored many strategies.  Among these, Few-Shot Learning (FSL) has become a research hotspot for solving the sample scarcity problem in HSI classification. FSL is different from traditional transfer learning. FSL aims to quickly adapt to new tasks using only a few labeled samples. This simulates how humans can rapidly learn new categories. Early FSL methods mainly focused on metric learning. Examples include Matching Networks \cite{Matching-net} and Prototypical Networks \cite{Prototypical_net}. These methods build a metric space. They then compare the similarity between a support set and a query set. This achieves few-shot classification. In the HSI field, researchers have combined these methods with HSI's spatio-spectral characteristics. For instance, Deep Few-Shot Learning (DFSL) \cite{dfsl} proposed a deep FSL method. It is based on 3D residual networks. DFSL learns HSI's spatio-spectral metric representations. This significantly improved classification performance.

As research progressed, cross-domain few-shot learning gained attention. Significant domain shifts exist between different HSI datasets. These shifts are due to factors like sensor differences and varying imaging environments. Therefore, directly transferring knowledge between different HSI datasets has limited effectiveness \cite{singh2020calibrating,tsengcross}. To address this, Li et al. \cite{dcfsl} proposed Deep Cross-Domain Few-Shot Learning (DCFSL). DCFSL combines domain adaptation with FSL. It uses adversarial learning. This reduces distribution differences between source and target domains. Zhang et al. \cite{giacfsl} introduced Graph Neural Networks (GNNs) to cross-domain FSL. They proposed an information aggregation framework. This framework builds graph structures for both source and target domains. It aggregates non-local spatial information. It also uses graph alignment strategies. These strategies achieve feature-level and distribution-level domain alignment. Ye et al. \cite{adafsl} used an adaptive strategy. This strategy assigns weights to conditional domain adversarial losses for different categories. This achieves global conditional distribution alignment. It effectively alleviates the problem of inconsistent domain distributions. To tackle source domain bias in cross-domain FSL, Qin et al. \cite{fdfsl} proposed the FDFSL method. This method is based on feature disentanglement. It decomposes features into domain-shared and domain-specific components. These components are then used specifically in meta-learning tasks for source and target domains. This approach suppresses the negative impact of source domain knowledge. It also helps to better learn target domain characteristics.

However, the high acquisition cost of HSI datasets limits the wide application of knowledge transfer across different HSI datasets. In contrast, natural image datasets have abundant labeled data. They also offer diverse feature representations. This makes them ideal source domains for heterogeneous knowledge transfer. Wang et al. pioneered the Heterogeneous FSL (HFSL) method \cite{hfsl}. HFSL pre-trains a spatial feature extractor on mini-ImageNet. It then fine-tunes the extractor using a few labeled HSI samples. This significantly improved HSI classification performance.

Recently, Self-Supervised Learning (SSL) has been introduced into the HSI few-shot classification field. SSL can learn powerful feature representations without labeled data \cite{ericsson2022self}. SSL designs pre-training tasks. These tasks mine intrinsic patterns from unlabeled data. This further reduces the reliance on labeled samples. For example, FSCF-SSL \cite{fscf-ssl} built upon HFSL. It introduced self-supervised learning. FSCF-SSL designed SSL tasks such as rotation prediction. This enhanced feature representation robustness. It also combined contrastive learning to mine spatio-spectral representations. This effectively addressed HFSL's limitations in feature modeling and data utilization.

\begin{figure}[H]
    \centering
    \includegraphics[width=0.75\columnwidth]{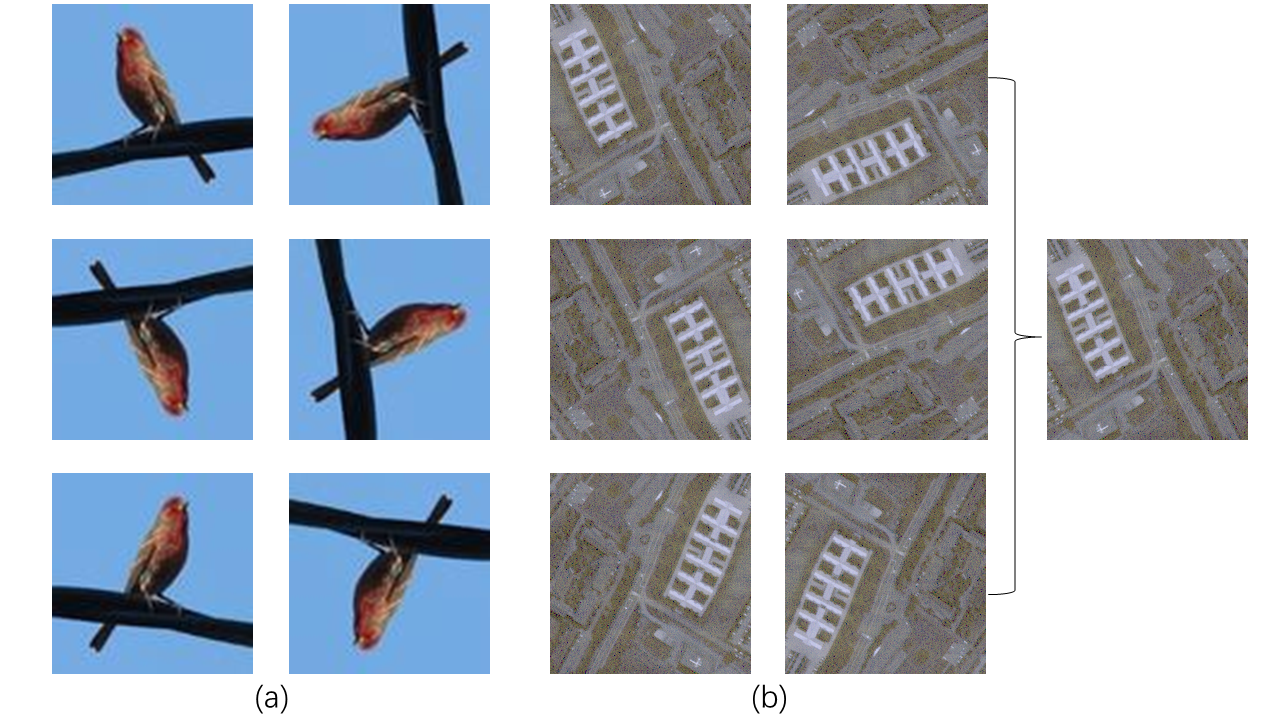}
    \caption{(a) Examples of typical geometric transformations applied to natural images.
(b) Examples of geometric transformations applied to the same HSI image patch.}
    \label{figure-bird-hsi}
\end{figure}

Existing methods have made significant progress in HSI few-shot classification. These methods use domain adaptation or heterogeneous knowledge transfer from natural images with self-supervised learning. However, these methods still face challenges in adapting to the unique spatial geometric diversity of HSI and effectively utilizing its spectral prior information.
Firstly, at the spatial feature level, HSI image patches exhibit more complex and diverse directional distributions in the spatial domain than natural images. As shown in Figure \ref{figure-bird-hsi}, in HSI classification, the label depends only on the category of the central pixel. Spatial transformations do not alter the central pixel's position or category, implying that geometrically transformed HSI patches remain in the same class. This is different from the relatively fixed directionality of objects in natural images. The spatial pattern of each pixel neighborhood in HSI shows a high degree of diversity, making it difficult to directly and efficiently apply the spatial feature extraction strategy migrated from natural images.

Furthermore, at the spectral feature level, HSIs have high spectral dimensionality and complex inter-band correlations. Directly relying on the limited labeled samples in the target domain for end-to-end training, the model is not only difficult to fully learn and utilize these complex spectral characteristics, but also prone to overfitting on a small number of samples due to the lack of sufficient spectral prior knowledge, thus limiting the model's HSI classification performance and generalization ability.

To address the challenges of model difficulty in adapting to HSI's spatial geometric diversity and the lack of spectral prior information, this paper proposes a Spectral-Spatial Self-Supervised Learning for Few-shot Hyperspectral Image Classification (S4L-FSC) method. Specifically, to enable spatial features learned from heterogeneous datasets to better adapt to the spatial geometric diversity of HSI image patches, we designed a Rotation-Mirror Self-Supervised Learning (RM-SSL) task. By setting rotation and mirror flipping as supervisory signals and integrating this with few-shot learning, this approach trains the spatial feature extraction module, aiming to drive the model to learn diverse spatial representations and transferable meta-knowledge from heterogeneous data, thereby enhancing its adaptability to HSI's spatial diversity. Secondly, to enrich the model's spectral prior information, we independently pre-trained the spectral feature extraction module on a homogeneous HSI dataset. By combining few-shot learning and Mask Reconstruction Self-Supervised learning (MR-SSL), the model is guided to learn the intrinsic structural information and dependencies of spectral data, thereby obtaining generalized spectral features. Experimental results demonstrate that the proposed S4L-FSC method significantly outperforms existing methods on four public HSI datasets. The main contributions of this paper are summarized as follows:

\begin{enumerate}
  \item Utilizing heterogeneous datasets, we design a Rotation-Mirror Self-Supervised Learning (RM-SSL) task. This task is combined with FSL methods to pre-train the spatial feature extraction module. This approach sets rotation and mirroring as supervisory signals and integrates with FSL tasks. It enables the model to learn diverse spatial representations and transferable meta-knowledge from heterogeneous data, thereby adapting to the spatial diversity of HSI.
  \item Utilizing homogeneous datasets, we design a Masked Reconstruction Self-Supervised Learning (MR-SSL) task. This task is combined with FSL methods to pre-train the spectral feature extraction module. This approach trains the model to recover the original spectrum from randomly masked spectral vectors, enabling it to infer spectral correlations. Furthermore, by integrating the MR-SSL task with FSL tasks, this method guides the model to learn feature representations for pixel-level classification, thereby equipping it with rich spectral prior information.
\end{enumerate}

The remainder of this paper is organized as follows. Section II briefly reviews related work. Section III details our proposed S4L-FSC method. Section IV presents experiments and analyses. Section V concludes the paper.

\section{RELATED WORK}

\subsection{Few-Shot Learning}
FSL aims to quickly adapt to new tasks via a small number of labeled samples, mirroring the human visual system's rapid learning capability in novel categories. Its core lies in extracting transferable knowledge from extensive source-domain data and and achieving efficient classification in the target domain using very few samples \cite{guo2020broader}. FSL generally employs a task-based training paradigm, creating multiple tasks that simulate few-shot scenarios \cite{wang2020generalizing}. Each task has a support set with N classes, each having K labeled samples, termed N-way K-shot. The model uses the support set for learning or adaptation. The query set contains new samples from these N classes and is used to evaluate the model's performance on the task.

Given the high cost of acquiring labeled samples in HSI classification, FSL is a promising solution for the label scarcity problem. Current FSL research mainly explores three directions:

(1) Model-agnostic meta-learning (MAML): Proposed by Chelsea Finn et al. in 2017, MAML optimizes model initialization to enable rapid adaptation to new tasks with minimal gradient updates \cite{maml}.

(2) Metric-based methods: These methods, including siamese networks \cite{siamesenet}, prototypical networks \cite{Prototypical_net}, and matching networks \cite{Matching-net}, learn an embedding space to measure sample similarities. They classify samples via distance or similarity metrics in this space, making them well-suited for few-shot learning.

(3) Memory-based methods: These methods use external memory units to store task information for quick recall, making them suitable for tasks with strong correlations or sequential dependencies \cite{santoro2016meta}.

By extracting and transferring knowledge across related tasks, FSL allows models to quickly adapt to new categories with minimal samples, offering an efficient solution for few-shot HSI classification.

\subsection{Self-Supervised Learning}
Self-Supervised Learning (SSL) enables the learning of useful feature representations by designing pretext tasks from unlabeled data, thereby avoiding reliance on large amounts of labeled data \cite{ssl-survey}. Its core principle is to utilize the inherent structure of the data itself as a supervisory signal. The application of SSL in HSI classification has garnered considerable attention in recent years, as it can effectively alleviate the problem of scarce labeled samples while simultaneously unveiling the data's latent representational capabilities.

Contrastive learning is a mainstream SSL approach that learns robust feature representations by maximizing the similarity between positive sample pairs and minimizing it between negative sample pairs. SimCLR \cite{chen2020simple} generates positive pairs through data augmentation and optimizes the feature space using a contrastive loss. SimSiam \cite{chen2021exploring} employs a Siamese network architecture, introducing a prediction head and a stop-gradient mechanism to effectively prevent representation collapse without relying on negative samples. In HSI classification, contrastive learning is often used to enhance the discriminability of spatio-spectral features. For instance, researchers have designed superpixel-based contrastive tasks to fully leverage the spatial neighborhood information in HSI \cite{guan2022spatial}.

Generative methods, which learn feature representations by generating data, constitute another significant branch of SSL. Autoencoders (AEs) \cite{hinton2006reducing} learn compressed representations by reconstructing input data. Generative Adversarial Networks (GANs) \cite{gan} generate realistic samples through adversarial training. Masked Autoencoders (MAEs) \cite{he2022masked} learn robust and semantically rich feature representations by randomly masking parts of the input data, such as image pixels or features, and then training the model to reconstruct the masked portions.

Another important category of SSL methods involves predictive tasks, which utilize the inherent geometric or contextual information within the data to generate supervisory signals. For example, Rotation Prediction \cite{gidaris2018unsupervised} trains a model to identify the rotation angle applied to an image. Jigsaw Puzzles \cite{noroozi2016unsupervised} or relative position prediction \cite{doersch2015unsupervised} require the model to understand the spatial arrangement or relative relationships of image patches. Such methods compel the model to learn the intrinsic structure and spatial information of images. Notably, Rotation Prediction has been employed in HSI few-shot classification, such as the SSLGT module in FSCF-SSL \cite{fscf-ssl}, which aims to learn features adaptable to orientation changes.

\section{PROPOSED S4L-FSC METHOD}
This section details the proposed S4L-FSC method. For clarity, we first define the key datasets and symbols used in this study. FSL involves heterogeneous data $D_{s_1}$, homogeneous data $D_{s_2}$, and target domain data $D_t$. In the S4L-FSC method, the heterogeneous data is the mini-ImageNet dataset, and the homogeneous data is the Chikusei dataset. The target domain data $D_t$ is divided into two parts: $D_l$ and $D_u$. $D_l$ contains a few labeled samples. $D_u$ consists of unlabeled samples for testing. These parts satisfy the condition $D_u \cup D_l = D_t$. In the experiments, $D_{l_1}$ is derived from $D_l$. This dataset $D_{l_1}$ is formed by applying noise-based data augmentation to $D_l$. This process increases the number of samples per class to 200.

The datasets $D_{s_1}$, $D_{s_2}$, and $D_t$ are used for training on their respective data. The symbols used in this chapter are shown in Table \ref{tab:notation}.

\begin{figure*}[!t]
  \centering
  \includegraphics[width=\textwidth]{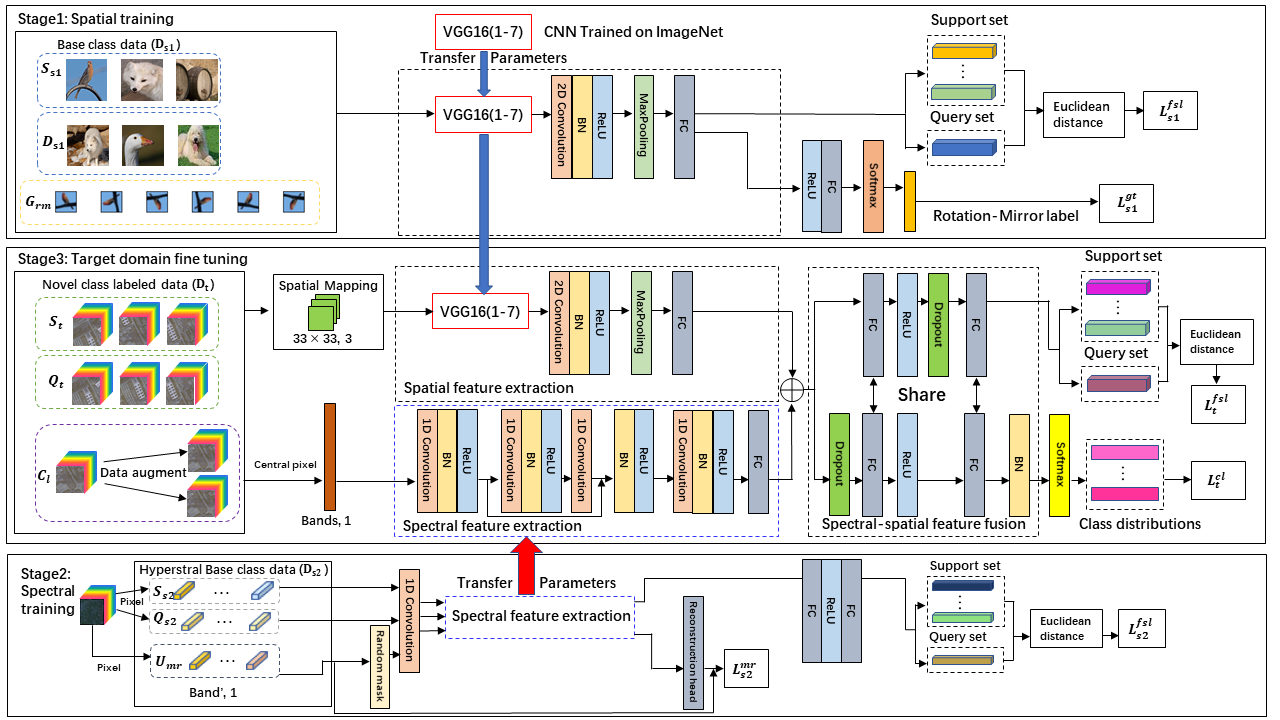} 
  \caption{Framework of the S4L-FSC Method. The training process involves three sequential stages: pre-training the spatial module on heterogeneous data by combining FSL with RM-SSL, pre-training the spectral module on homogeneous data by combining FSL with MR-SSL, and loading the parameters of both pre-trained modules for few-shot fine-tuning and classification on the target domain by integrating FSL with contrastive learning.}
  \label{figure-framework2}
\end{figure*}

\subsection{Overview of the S4L-FSC Method}
The overall framework of the S4L-FSC method is shown in Figure~\ref{figure-framework2}. It includes three stages.
The first stage is heterogeneous data pre-training. This pre-training is conducted on the Mini-ImageNet dataset. It combines FSL and RM-SSL to train the spatial feature extraction module.
The second stage is homogeneous data pre-training. This occurs on the Chikusei dataset. It combines FSL and MR-SSL to train the spectral feature extraction module.
The third stage is target domain few-shot fine-tuning. This is performed on the target domain dataset. In this stage, the pre-trained spectral and spatial modules are loaded. The method then combines FSL and contrastive learning for fine-tuning. Finally, testing is conducted.

\begin{table*}[!htbp]
	
    \centering

    \caption{Main symbols and definitions used in S4L-FSC}

    \label{tab:notation}

    \small 

    \begin{tabular}{ccc}

        \toprule 

        Symbol & Definition & Notes \\

        \hline


        $D_{s_1}$ & Heterogeneous dataset & \multirow{14}{*}{Stage 1} \\

        $S_{s_1}$ & Support set of $D_{s_1}$ & \\

        $Q_{s_1}$ & Query set of $D_{s_1}$ & \\

        $G_{\text{rm}}$ & Transformed sample set for RM-SSL & \\

        $N_{s_1}$ & Number of classes for FSL on $D_{s_1}$ & \\

        $c_{m}^{\text{spatial}}$ & Class prototype & \\

        $x_{s_1}^{\text{sup},i}, y_{s_1}^{\text{sup},i}$ & The $i$-th sample and label in $S_{s_1}$ & \\

        $x_{s_1}^{\text{que},j}, y_{s_1}^{\text{que},j}$ & The $j$-th sample and label in $Q_{s_1}$ & \\

        $L_{s_1}^{\text{fsl}}$ & FSL loss on $D_{s_1}$ & \\

        $L_{s_1}^{\text{rm}}$ & RM-SSL loss & \\

        $L_{s_1}^{\text{total}}$ & Total loss for Stage 1 & \\

        $F_{\text{spatial}}(\cdot; \theta_{\text{spatial}})$ & Spatial feature extraction module and parameters & \\

        $T_{\text{rm}}$ & Set of geometric operations for RM-SSL & \\

        $R_{\text{rm}}(\cdot; \phi)$ & RM-SSL classification head and parameters & \\

        \hline


        $D_{s_2}$ & Homogeneous dataset & \multirow{18}{*}{Stage 2} \\

        $S_{s_2}$ & Support set of $D_{s_2}$ & \\

        $Q_{s_2}$ & Query set of $D_{s_2}$ & \\

        $M_{\text{spec}}(\cdot; \omega_{\text{map}})$ & Spectral mapping layer and parameters & \\

        $F_{\text{spectral}}(\cdot; \theta_{\text{spectral}})$ & Spectral feature extraction module and parameters & \\

        $H_{\text{FSL}}(\cdot; \theta_{\text{linear}})$ & FSL linear layer and parameters & \\

        $c_{m}^{\text{spectral}}$ & Class prototype & \\

        $x_{s_2}^{\text{sup},i}, y_{s_2}^{\text{sup},i}$ & The $i$-th sample and label in $S_{s_2}$ & \\

        $x_{s_2}^{\text{que},j}, y_{s_2}^{\text{que},j}$ & The $j$-th sample and label in $Q_{s_2}$ & \\

        $L_{s_2}^{\text{fsl}}$ & FSL loss on $D_{s_2}$ & \\

        $L_{s_2}^{\text{mr}}$ & MR-SSL loss & \\

        $L_{s_2}^{\text{total}}$ & Total loss for Stage 2 & \\

        $x_{s_2} \in \mathbb{R}^{B_{s_2}}$ & Original spectral sample for MR-SSL & \\

        $\tilde{x}_{s_2}$ & Masked spectral sample for MR-SSL & \\

        $z_{s_2}^{\text{mr}}$ & Latent representation in MR-SSL & \\

        $D_{\text{recon}}(\cdot; \psi)$ & MR-SSL decoder and parameters & \\

        $\hat{x}_{s_2}$ & Reconstructed spectral vector in MR-SSL & \\

        $B_{s_2}$ & Original spectral dimension of $D_{s_2}$ & \\

        \hline


        $D_{t}$ & Target domain dataset & \multirow{13}{*}{Stage 3} \\ 

        $D_{l}$ & Target domain labeled dataset & \\

        $D_{u}$ & Target domain unlabeled test set & \\

        $D_{l_1}$ & Augmented target domain labeled set & \\

        $F_{\text{fused}}(\cdot; \eta)$ & Spectral-spatial feature extraction network and parameters & \\

        $S_{t}$ & Support set of $D_t$ & \\

        $Q_{t}$ & Query set of $D_t$ & \\

        $L_{t}^{\text{fsl}}$ & FSL loss on $D_t$ & \\

        $c_{m}^{\text{fused}}$ & Class prototype & \\

        $\mathcal{T}(\cdot)$ & Data augmentation function for SSLCL & \\

        $(x^{\mathcal{A}_1}, x^{\mathcal{A}_2})$ & Augmented sample pair for SSLCL & \\

        $L_{t}^{\text{cl}}$ & SSLCL loss & \\

        $N_{t}$ & Number of classes for FSL on $D_t$ & \\

        \bottomrule

    \end{tabular}

\end{table*}

On the heterogeneous dataset $D_{s_1}$, samples are randomly selected to form a support set and a query set. The model is then trained using a metric-based meta-learning approach to implement FSL. For RM-SSL, samples are randomly chosen from $D_{s_1}$ to construct rotation-mirror samples $G_{\text{rm}}$ for model training. Both FSL and RM-SSL are performed on this heterogeneous data to learn transferable meta-knowledge and diverse spatial representations. The total loss for the heterogeneous data is:
\begin{equation}
    L_{s_1}^{\text{total}} = L_{s_1}^{\text{fsl}} + L_{s_1}^{\text{rm}}
\end{equation}

For FSL on the homogeneous dataset $D_{s_2}$, samples are randomly selected. Other FSL settings are identical to those used for the heterogeneous dataset. For MR-SSL on this homogeneous data, samples are also randomly chosen from $D_{s_2}$ for training. This process encourages the model to capture spectral dependencies and patterns, thereby equipping it with rich spectral prior information. The total loss on the homogeneous dataset is:
\begin{equation}
    L_{s_2}^{\text{total}} = L_{s_2}^{\text{fsl}} + L_{s_2}^{\text{mr}}
\end{equation}

During the target domain few-shot fine-tuning process, the feature extraction model employed is a spectral-spatial feature extraction network. The parameters for this network's spatial feature extraction module are transferred from the model pre-trained on heterogeneous data. Similarly, parameters for its spectral feature extraction module are transferred from the model pre-trained on homogeneous data.
For Self-Supervised Learning with Contrastive Learning (SSLCL) on the target domain data, $D_l$ serves as the input. FSL and SSLCL are performed concurrently on the target domain data. This enables the model to learn more discriminative features. The total loss on the target domain dataset is:
\begin{equation}
    L_{t}^{\text{total}} = L_{t}^{\text{fsl}} + L_{t}^{\text{cl}}
\end{equation}

\subsection{Network Architecture}

To effectively extract joint spatio-spectral features from hyperspectral images (HSI), this study adopts the network structure proposed in FSCF-SSL \cite{fscf-ssl}. This network consists of three main components:

\begin{enumerate}[leftmargin=0pt, labelwidth=18pt, itemindent=18pt, label=\textit{\arabic*)}]
    \item Spatial Feature Extraction Module: This module draws inspiration from the VGG16 architecture. It utilizes the weights of the first seven layers of VGG16 pre-trained on the ImageNet dataset. To process HSI data with dimensions $33 \times 33 \times B$, the HSI data is first mapped to three channels. This is done using a mapping layer composed of a $1 \times 1$ convolution. The data dimensions are thus reshaped to $33 \times 33 \times 3$. Subsequently, spatial feature vectors are extracted after the data passes through the VGG backbone network and additional convolutional and fully connected layers.

    \item Spectral Feature Extraction Module: This module processes the spectral vector input from the central pixel. It employs a 1D Convolutional Neural Network. The module includes an initial 1D convolutional layer. This is followed by a spectral residual block. The spectral residual block contains two 1D convolutional layers and an identity mapping. Finally, spectral features are output through a fully connected layer. This design aims to capture fine-grained spectral structural information.

    \item Spectral-Spatial Feature Fusion Module: Spatial features and spectral features are concatenated to obtain fused features. These fused features then pass through shared fully connected layers. To accommodate the different requirements of FSL and SSLCL tasks, the network layers in the fusion module differ. For the FSL task, the layers are sequentially: a fully connected layer, ReLU, Dropout, and another fully connected layer. For the SSLCL task, the layers are sequentially: Dropout, a fully connected layer, ReLU, another fully connected layer, a BatchNorm layer, and a Softmax layer.
\end{enumerate}

The detailed parameters of the network architecture are presented in Table \ref{tab:network_params}.

\begin{table*}[!t]
  \centering
  \caption{Network Architecture and Parameters}
  \label{tab:network_params}
  \begin{tabular}{ccccccc}
    \toprule
    \textbf{Module} & \textbf{Type} & \textbf{Input Shape} & \textbf{Output Shape} & \textbf{Kernel Size} & \textbf{Stride} & \textbf{Padding} \\
    \midrule
    \multirow{2}{*}{Spatial Mapping} 
    & Conv2d & (n\_bands, 33, 33) & (3, 33, 33) & $1\times1$ & 1 & 0 \\
    & BatchNorm2d & (3, 33, 33) & (3, 33, 33) & -- & -- & -- \\
    \midrule
    \multirow{5}{*}{Spatial Feature Extraction} 
    & VGG16(1-7) & (3, 33, 33) & (256, 4, 4) & -- & -- & -- \\
    & Conv2d & (256, 4, 4) & (512, 2, 2) & $3\times3$ & 1 & 0 \\
    & BatchNorm2d & (512, 2, 2) & (512, 2, 2) & -- & -- & -- \\
    & MaxPool2d & (512, 2, 2) & (512, 1, 1) & $2\times2$ & 2 & -- \\
    & Linear & 512 & 100 & -- & -- & -- \\
    \midrule
    \multirow{9}{*}{Spectral Feature Extraction} 
    & Conv1d & (1, n\_bands) & (24, $\lfloor$(n\_bands - 7 + 1)/2$\rfloor$) & 7 & 2 & 0 \\
    & BatchNorm1d & (24, $\lfloor$(n\_bands - 7 + 1)/2$\rfloor$) & (24, $\lfloor$(n\_bands - 7 + 1)/2$\rfloor$) & -- & -- & -- \\
    & Conv1d & (24, $\lfloor$(n\_bands - 7 + 1)/2$\rfloor$) & (24, $\lfloor$(n\_bands - 7 + 1)/2$\rfloor$) & 7 & 1 & 3 \\
    & BatchNorm1d & (24, $\lfloor$(n\_bands - 7 + 1)/2$\rfloor$) & (24, $\lfloor$(n\_bands - 7 + 1)/2$\rfloor$) & -- & -- & -- \\
    & Conv1d & (24, $\lfloor$(n\_bands - 7 + 1)/2$\rfloor$) & (24, $\lfloor$(n\_bands - 7 + 1)/2$\rfloor$) & 7 & 1 & 3 \\
    & BatchNorm1d & (24, $\lfloor$(n\_bands - 7 + 1)/2$\rfloor$) & (24, $\lfloor$(n\_bands - 7 + 1)/2$\rfloor$) & -- & -- & -- \\
    & Conv1d & (24, $\lfloor$(n\_bands - 7 + 1)/2$\rfloor$) & (128, 1) & $\lfloor$(n\_bands - 7)/2$\rfloor$ + 1 & 1 & 0 \\
    & BatchNorm1d & (128, 1) & (128, 1) & -- & -- & -- \\
    & Linear & 128 & 100 & -- & -- & -- \\
    \midrule
    \multirow{3}{*}{Spectral-Spatial Feature Fusion} 
    & Linear & 200 & 64 & -- & -- & -- \\
    & Dropout & 64 & 64 & -- & -- & -- \\
    & Linear & 64 & n\_classes & -- & -- & -- \\
    \bottomrule
  \end{tabular}
\end{table*}

\subsection{Few-Shot Learning on Heterogeneous Data}
In the first stage of training, FSL is applied to the heterogeneous dataset $D_{s_1}$. The objective is to enable the spatial feature extraction module to learn transferable meta-knowledge. In each episode, $N_{s_1}$ classes are randomly sampled from $D_{s_1}$. For each class, $K$ samples form the support set $S_{s_1} = \{ ( x_{s_1}^{\text{sup},i}, y_{s_1}^{\text{sup},i} ) \}_{i=1}^{K \times N_{s_1}}$, and $C$ samples form the query set $Q_{s_1} = \{ ( x_{s_1}^{\text{que},j}, y_{s_1}^{\text{que},j} ) \}_{j=1}^{C \times N_{s_1}}$.

For the support set $S_{s_1}$, features are extracted using the spatial feature extractor $F_{\text{spatial}}(\cdot; \theta_{\text{spatial}})$. Then, the prototype $c_{m}^{\text{spatial}}$ for each class $m$ is calculated. This prototype is the mean of the features from all support set samples belonging to that class.For a query sample $x_{s_1}^{\text{que},j}$ from the query set, its feature $F_{\text{spatial}}(x_{s_1}^{\text{que},j})$ is first computed. Next, the Euclidean distance $d(\cdot, \cdot)$ between this feature and each class prototype $c_{m}^{\text{spatial}}$ is calculated. These distance values are then converted into a probability distribution of the sample belonging to class $m$ using the Softmax function. The formula is as follows:
\begin{equation}
p(y_{s_1}^{\text{que},j}=m | x_{s_1}^{\text{que},j}) = \frac{\exp(-d(F_{\text{spatial}}(x_{s_1}^{\text{que},j}), c_m^{\text{spatial}}))}{\sum_{m'=1}^{N_{s_1}} \exp(-d(F_{\text{spatial}}(x_{s_1}^{\text{que},j}), c_{m'}^{\text{spatial}}))} \label{eq:prob_s1}
\end{equation}

The FSL loss for this stage, $L_{s_1}^{\text{fsl}}$, is calculated by minimizing the negative log-likelihood of the query set, as shown in Equation~\ref{eq:fsl_s1}.
\begin{equation}
    L_{s_1}^{\text{fsl}} = \mathbb{E}_{S_{s_1}, Q_{s_1} \sim D_{s_1}} \left[ -\sum_{j=1}^{C \times N_{s_1}} \log p(y_{s_1}^{\text{que},j} | x_{s_1}^{\text{que},j}) \right]
    \label{eq:fsl_s1}
\end{equation}

By optimizing the FSL loss of heterogeneous data, the model is encouraged to learn how to more effectively cluster similar samples in the vicinity of the corresponding prototype in the feature space, thereby improving the classification performance.

\subsection{Rotation-Mirror Self-Supervised Learning}

The Rotation-Mirror Self-Supervised Learning (RM-SSL) task aims to enhance the adaptability of the spatial feature extraction module to the spatial geometric diversity of HSI. It achieves this by training the module to predict geometric transformations of images.

The task first defines a set of six geometric transformation operations, $T_{\text{rm}} = \{t_k | k=1, \dots, 6\}$. Specifically, $t_1, t_2, t_3, \text{and } t_4$ represent rotations of the original image by $0^\circ, 90^\circ, 180^\circ, \text{and } 270^\circ$, respectively. $t_5$ represents a horizontal flip of the original image. $t_6$ represents a vertical flip of the original image.

For each sample $x_{s_1}$ drawn from the heterogeneous dataset $D_{s_1}$, all six transformations in $T_{\text{rm}}$ are applied. This generates a set of transformed images: $\{ x_{s_1}^{(k)} = t_k(x_{s_1}) | k=1, \dots, 6 \}$. The objective of the RM-SSL task is to train a network. This network consists of the spatial feature extractor $F_{\text{spatial}}$ followed by a classification network $R_{\text{rm}}(\cdot; \phi)$. This combined network predicts the transformation class label $k$ that was applied to the image $x_{s_1}^{(k)}$. The training data for this task is represented as $G_{\text{rm}} = \{ (x_{s_1}^{(k)}, k) | x_{s_1} \in D_{s_1}, k \in \{1, \dots, 6\} \}$.

The corresponding self-supervised loss function, $L_{s_1}^{\text{rm}}$, uses the cross-entropy loss form. It is defined as follows:
\begin{equation}
L_{s_1}^{\text{rm}} = \mathbb{E}_{(x_{s_1}^{(k)}, k) \sim G_{\text{rm}}} \left[ -\log p(k | x_{s_1}^{(k)}; \theta_{\text{spatial}}, \phi) \right]
\label{eq:rm_ssl_loss} 
\end{equation}
Here, $p(k|x_{s_1}^{(k)})$ is the probability that the classification head $R_{\text{rm}}(F_{\text{spatial}}(x_{s_1}^{(k)}; \theta_{\text{spatial}}); \phi)$ predicts that image $x_{s_1}^{(k)}$ belongs to transformation class $k$. The terms $\theta_{\text{spatial}}$ and $\phi$ are the learnable parameters of the spatial feature extractor and the RM-SSL classification head, respectively.

By learning to predict these geometric transformations, the model can capture image features under different orientations and perspectives. This, in turn, improves its adaptability to the spatial geometric diversity of HSI.

\subsection{Few-Shot Learning on Homogeneous Data}

In the second training stage, FSL is applied to the homogeneous dataset $D_{s_2}$. This dataset contains abundant HSI labeled samples. The purpose is to pre-train the spectral feature extraction module, $F_{\text{spectral}}(\cdot; \theta_{\text{spectral}})$. This pre-training helps the module learn feature representations for pixel-level classification.
In each episode, $N_{s_2}$ classes are randomly sampled from $D_{s_2}$. These samples form a support set $S_{s_2}$ and a query set $Q_{s_2}$. Samples from the support and query sets first pass through a 1D convolutional mapping layer, $M_{\text{spec}}(\cdot; \omega_{\text{map}})$. This layer adjusts their dimensionality. The goal is to align the spectral dimensions of the homogeneous dataset with the target domain dataset. After this, the samples pass through the spectral feature extraction module $F_{\text{spectral}}(\cdot; \theta_{\text{spectral}})$ and a linear layer $H_{\text{FSL}}(\cdot; \theta_{\text{linear}})$. This process yields feature representations suitable for metric-based comparison.
The prototype $c_{m}^{\text{spectral}}$ for each class $m$ is calculated based on the feature representations of the support set samples. The classification probability for a query sample $x_{s_2}^{\text{que},j}$ is then determined. This determination is based on the Euclidean distance between its final feature representation and each class prototype $c_{m}^{\text{spectral}}$.

The corresponding FSL loss, $L_{s_2}^{\text{fsl}}$, aims to minimize the classification error on the query set $Q_{s_2}$. It is calculated as shown in Equation~\ref{eq:fsl_s2}.
\begin{equation}
\begin{split}
L_{s_2}^{\text{fsl}} ={} & \mathbb{E}_{S_{s_2}, Q_{s_2} \sim D_{s_2}} \left[ -\sum_{j=1}^{C \times N_{s_2}} \log p(y_{s_2}^{\text{que},j} | x_{s_2}^{\text{que},j}; \right. \\
& \left. \theta_{\text{spectral}}, \omega_{\text{map}}, \theta_{\text{linear}}) \right]
\end{split}
\label{eq:fsl_s2}
\end{equation}
Here, the calculation of the probability $p(y_{s_2}^{\text{que},j} | x_{s_2}^{\text{que},j})$ depends on two factors. The first is the features obtained using parameters $\theta_{\text{spectral}}, \omega_{\text{map}}, \text{and } \theta_{\text{linear}}$. The second is the corresponding prototype $c_{m}^{\text{spectral}}$.

Optimizing the FSL loss on homogeneous data minimizes the classification error on the query set. This helps the spectral feature module acquire HSI prior information. Consequently, it improves the model's ability to distinguish between different classes.

\subsection{Masked Reconstruction Self-Supervised Learning}

To enable the spectral feature extractor $F_{\text{spectral}}$ to infer spectral correlations, a masked reconstruction self-supervised learning task is introduced in the second training stage. MR-SSL utilizes the homogeneous HSI dataset $D_{s_2}$. It compels the model to infer and understand spectral correlations by reconstructing the original spectrum from randomly masked inputs.

Specifically, consider a given spectral vector $x_{s_2} \in \mathbb{R}^{B}$ from $D_{s_2}$. First, a mask vector $\bm{mask} \in \{0, 1\}^B$ is randomly generated. Here, $mask_i = 1$ indicates that the $i$-th band is masked, while $mask_i = 0$ indicates it is visible. The proportion of masked bands is determined by a predefined ratio. This ratio controls the difficulty of the masking task and the intensity of the self-supervised learning. The masked input $\tilde{x}_{s_2}$ is obtained by setting the band values at the masked positions to zero:\begin{equation}
    \tilde{x}_{s_2} = x_{s_2} \odot (1-\bm{mask})
\end{equation}
where $\odot$ denotes the element-wise Hadamard product.

Subsequently, the masked spectrum $\tilde{x}_{s_2}$ is first fed into a 1D convolutional mapping layer, $M_{\text{spec}}(\cdot; \omega_{\text{map}})$, for dimensionality adjustment. Next, it is processed by the spectral feature extractor, $F_{\text{spectral}}(\cdot; \theta_{\text{spectral}})$, to obtain the latent representation $z_{s_2}^{\text{mr}}$. Finally, a dedicated decoder network, $D_{\text{recon}}(\cdot; \psi)$, takes this latent representation $z_{s_2}^{\text{mr}}$ as input to reconstruct the original, complete spectral vector $\hat{x}_{s_2}$:
\begin{equation}
    \hat{x}_{s_2} = D_{\text{recon}}(z_{s_2}^{\text{mr}}; \psi)
\end{equation}
Here, $\theta_{\text{spectral}}$ and $\psi$ are the learnable parameters of the spectral encoder and the reconstruction decoder, respectively.

The optimization objective of MR-SSL is to minimize the reconstruction error between the original spectrum $x_{s_2}$ and the reconstructed spectrum $\hat{x}_{s_2}$. In the implementation of MR-SSL, the Mean Squared Error (MSE) is calculated across the entire spectral dimension. This measures the difference between the reconstructed and original spectra over all $B$ bands. Therefore, the MR-SSL loss, $L_{s_2}^{\text{mr}}$, is defined as:
\begin{equation}
L_{s_2}^{\text{mr}} = \mathbb{E}_{x_{s_2} \sim D_{s_2}, \bm{mask}} \left[ \frac{1}{B} \sum_{i=1}^B (x_{s_2, i} - \hat{x}_{s_2, i})^2 \right]
\label{eq:mr_ssl_loss_full}
\end{equation}

In this equation, $x_{s_2, i}$ and $\hat{x}_{s_2, i}$ represent the values of the $i$-th band in the original and reconstructed vectors, respectively. $B$ is the total number of spectral bands. The expectation $\mathbb{E}$ is computed over the data distribution $D_{s_2}$ and the random mask $\bm{mask}$. Minimizing this loss encourages the spectral encoder and decoder to effectively recover the complete spectral information from the randomly masked input. This equips the model with rich spectral prior information, thereby enhancing its classification performance.

\subsection{Target Domain Few-Shot Learning}

In target domain FSL, the pre-trained spatial feature extractor $F_{\text{spatial}}$ and spectral feature extractor $F_{\text{spectral}}$ are first loaded. These two modules are integrated via a feature fusion module. This forms the spectral-spatial feature extraction network used for target domain classification. The spectral-spatial feature extraction network model is denoted as $F_{\text{fused}}(\cdot; \eta)$. Here, $\eta$ includes the optimizable parameters from $\theta_{\text{spatial}}$, $\theta_{\text{spectral}}$, and the spectral-spatial feature fusion module.

To quickly adapt to target categories using a small number of labeled samples $D_l$, the dataset $D_{l_1}$, generated via data augmentation, is used in this stage. During each iteration, a support set $S_t = \{ (x_{l_1}^{\text{sup}, i}, y_{l_1}^{\text{sup}, i}) \}_{i=1}^{K \times N_t}$ with $K$ support samples per class and a query set $Q_t = \{ (x_{l_1}^{\text{que}, j}, y_{l_1}^{\text{que}, j}) \}_{j=1}^{C \times N_t}$ with $C$ query samples per class are randomly sampled from $D_{l_1}$. The training loss, $L_t^{\text{fsl}}$, is defined as the expected negative log-likelihood of the query set in each training episode:
\begin{equation}
    L_t^{\text{fsl}} = \mathbb{E}_{S_t, Q_t \sim D_{l_1}} \left[ -\sum_{j=1}^{C \times N_t} \log p(y_{l_1}^{\text{que},j} | x_{l_1}^{\text{que},j}; \eta) \right]
    \label{eq:fsl_target_loss}
\end{equation}
where $p(y_{l_1}^{\text{que},j} | x_{l_1}^{\text{que},j}; \eta)$ represents the classification probability of the query sample $x_{l_1}^{\text{que},j}$. This probability is obtained by first calculating the prototype $c_m^{\text{fused}}$ for each class from the support set $S_t$. Then, the Euclidean distance between the features of the query sample $x_{l_1}^{\text{que},j}$ and each prototype $c_m^{\text{fused}}$ is computed. Finally, these distances are converted into probabilities using the Softmax function.

\subsection{Self-Supervised Learning With Contrastive Learning}

The purpose of Self-Supervised Learning with Contrastive Learning (SSLCL) is to further leverage the limited labeled samples $D_l$ during the target domain few-shot fine-tuning stage. This aims to mine their intrinsic structural information and enhance the discriminative ability of the spectral-spatial feature extraction network $F_{\text{fused}}$.

Unlike target domain FSL, which uses the augmented dataset $D_{l_1}$, the SSLCL task operates directly on the original set of few labeled samples $D_l$ from the target domain. The sample set $D_l$ contains $N_t$ classes, with $K_0$ labeled samples per class. In each training iteration, the SSLCL task uses all $N_t \times K_0$ samples from $D_l$ as its input batch. Thus, the batch size is $B_s = N_t \times K_0$.

For each original sample $x_l$ in batch $B_s$, two augmented views, $x^{\mathcal{A}_1} = \mathcal{T}(x_l)$ and $x^{\mathcal{A}_2} = \mathcal{T}(x_l)$, are generated. This is achieved by independently applying a random data augmentation function $\mathcal{T}(\cdot)$ twice. The random data augmentation function $\mathcal{T}(\cdot)$ includes a series of random transformations for HSI patches. These mainly include: (1) applying random crop or Gaussian noise injection with a certain probability; (2) random horizontal flip; (3) random vertical flip; and (4) random rotation by multiples of $90^\circ$. Specific details of transformation combinations and hyperparameter settings are provided in the experimental setup section.

These two augmented views, $x^{\mathcal{A}_1}$ and $x^{\mathcal{A}_2}$, are then passed through the spectral-spatial feature extraction network $F_{\text{fused}}(\cdot; \eta)$. This process outputs their respective class probability distributions. We denote these output class probability distributions for the two views as $z_t^{\text{cl},1}$ and $z_t^{\text{cl},2}$, respectively.

To encourage the model to learn class-invariant features, this paper adopts a self-supervised loss function originating from the literature \cite{wang2023self}. This function optimizes the consistency of outputs between different data views. Combined with the effects of sharpness and diversity terms, it aims to prevent model representation collapse and enhance feature discriminability.
This loss, $L_t^{\text{cl}}$, is defined as follows:
\begin{equation}
    L_t^{\text{cl}}(z_t^{\text{cl},1}, z_t^{\text{cl},2}; \eta) = \frac{1}{2} \left( L(z_t^{\text{cl},1} || z_t^{\text{cl},2}) + L(z_t^{\text{cl},2} || z_t^{\text{cl},1}) \right)
    \label{eq:sslcl_main}
\end{equation}
where $L(z_t^{\text{cl},1} || z_t^{\text{cl},2})$ is calculated as:
\begin{equation}
\label{eq:sslcl_sub} 
\begin{split}
    & L(z_t^{\text{cl},1} || z_t^{\text{cl},2}) = \quad \underbrace{\frac{1}{B_s} \sum_{i=1}^{B_s} D_{\text{KL}}(z_t^{\text{cl}, i, 1} || z_t^{\text{cl}, i, 2})}_{\text{consistency term}}  \\ 
    &  + \underbrace{\left( \frac{1}{B_s} \sum_{i=1}^{B_s} H(z_t^{\text{cl}, i, 1}) \right)}_{\text{sharpness term}} - \underbrace{H\left( \frac{1}{B_s} \sum_{i=1}^{B_s} z_t^{\text{cl}, i, 1} \right)}_{\text{diversity term}} 
\end{split}
\end{equation}
Here, $z_t^{\text{cl}, i, 1}$ is the output probability distribution of the first view for the $i$-th sample, and $z_t^{\text{cl}, i, 2}$ is that of its second view. $D_{\text{KL}}(\cdot||\cdot)$ represents KL divergence, and $H(\cdot)$ represents Shannon entropy.

\section{EXPERIMENTAL RESULTS AND ANALYSIS}

\subsection{Experimental Dataset}
To comprehensively evaluate the performance of the S4L-FSC method, we used four hyperspectral datasets in our experiments. Specifically, Mini-ImageNet was used as the heterogeneous dataset $D_{s_1}$. The Chikusei dataset was used as the homogeneous dataset $D_{s_2}$. We also selected four widely-used HSI datasets as target domain datasets $D_t$ for few-shot classification assessment. These are: University of Pavia (UP), Indian Pines (IP), Salinas (SA), and WHU-Hi-HanChuan (HC).

\subsubsection{Heterogeneous Source Dataset}
The Mini-ImageNet dataset is a subset of ImageNet. It contains 100 classes. Each class has 600 color images of $84 \times 84$ pixels. In the S4L-FSC method, this dataset serves as heterogeneous data. It is used for pre-training the spatial feature extraction module.

\subsubsection{Homogeneous Source Dataset}
\begin{figure}[!htbp] 
\small
\captionsetup{font=small} 
  \centering
  \includegraphics[width=1.05\linewidth]{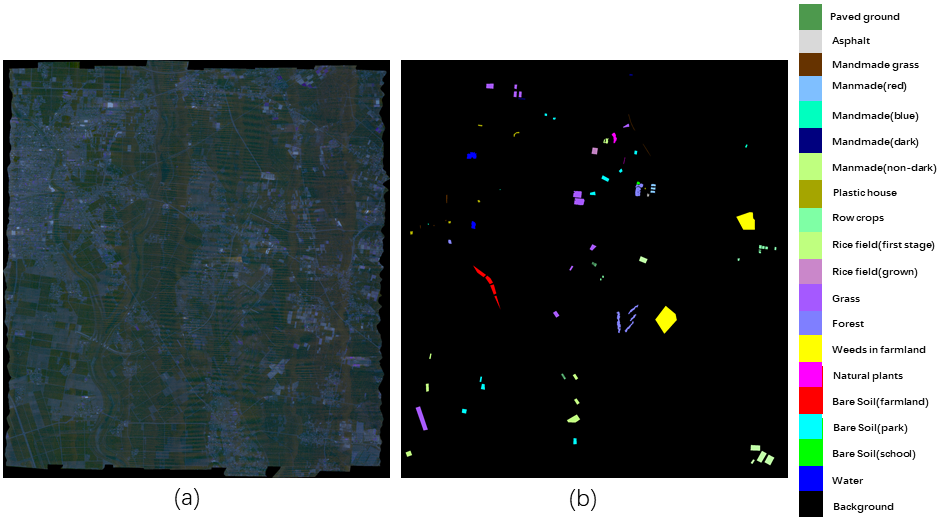} 
  \caption{Chikusei dataset: (a) Pseudocolor image,(b) Ground-truth map.}
  \label{fig:Chikusei}
\end{figure}

The Chikusei dataset is an airborne hyperspectral dataset. It was acquired by the Headwall Hyperspec-VNIR-C sensor over Chikusei, Japan, in July 2014. This dataset covers the spectral range from 363~nm to 1018~nm. It contains a total of 128 spectral bands. The image size is $2517 \times 2335$ pixels, with a spatial resolution of 2.5 meters. The dataset includes 19 annotated land cover classes, encompassing urban areas and farmland. In the S4L-FSC method, this dataset is used as homogeneous data. It serves for pre-training the spectral feature extraction module. Table~\ref{tab-Chikusei} shows the land cover classes and sample counts for the Chikusei dataset. Figure~\ref{fig:Chikusei} displays its pseudo-color image and ground truth map.

\begin{table}[!htbp] 
  \small 
  \caption{Class Name and Corresponding Sample Number in the Chikusei Dataset} 
  \label{tab-Chikusei} 
  \centering
  \begin{tabularx}{\linewidth}{c >{\raggedright\arraybackslash}X c} 
  \toprule
  \textbf{Class} & \textbf{Name} & \textbf{Number of Samples} \\
  \midrule
  1 & Water & 2845 \\
  2 & Bare Soil (school) & 2859 \\
  3 & Bare Soil (park) & 286 \\
  4 & Bare Soil (farmland) & 48525 \\
  5 & Natural plants & 4297 \\
  6 & Weeds in farmland & 1108 \\
  7 & Forest & 20516 \\
  8 & Grass & 6515 \\
  9 & Rice field (grown) & 13369 \\
  10 & Rice field (first stage) & 1268 \\
  11 & Row crops & 5961 \\
  12 & Plastic house & 2193 \\
  13 & Manmade (non-dark) & 1220 \\
  14 & Manmade (dark) & 7664 \\
  15 & Manmade (blue) & 431 \\
  16 & Manmade (red) & 222 \\
  17 & Manmade grass & 1040 \\
  18 & Asphalt & 801 \\
  19 & Paved ground & 145 \\
  \midrule
  \textbf{Total} & & \textbf{77592} \\ 
  \bottomrule
  \end{tabularx}%
\end{table}

\subsubsection{Target Domain Datasets}
The following four HSI datasets are used as target domain data. They serve to evaluate the model's classification performance in few-shot scenarios:

\begin{itemize}
    \item UP dataset: This dataset was acquired by the ROSIS sensor over the University of Pavia, Italy. The original data included 115 bands. After removing 12 noisy bands, 103 bands (0.43--0.86~$\mu\text{m}$) were retained for experiments. The image size is $610 \times 340$ pixels. Its spatial resolution is 1.3~meters/pixel. It contains a total of 9 land cover classes.
    \item IP dataset: This dataset was acquired by the AVIRIS sensor in 1992. The location was the Indian Pines test site in Northwestern Indiana, USA. The original data comprised 224 bands (0.4--2.5~$\mu\text{m}$). After removing water absorption and noisy bands, 200 bands were kept for experiments. The image size is $145 \times 145$ pixels, with a spatial resolution of 20~meters/pixel. The dataset contains 16 land cover classes.
    \item SA dataset: This dataset was also acquired by the AVIRIS sensor. It was collected over Salinas Valley, California, USA. The spectral range is 0.4--2.5~$\mu\text{m}$. After removing water absorption bands, 204 bands were retained. The image size is $512 \times 217$ pixels. Its spatial resolution is 3.7~meters/pixel. It includes 16 land cover classes.
    \item HC dataset: This dataset was acquired using a Headwall Nano-Hyperspec sensor mounted on a UAV platform. The image size is $1217 \times 303$ pixels, containing 274 bands. Its spatial resolution is approximately 0.109~meters. The dataset includes 16 land cover classes. Notably, due to the large total number of samples in this dataset, we randomly selected only 15\% of the samples from each class for our experiments.
\end{itemize}

\begin{figure}[!htbp]
  \centering
  \small
  \includegraphics[width=0.7\linewidth]{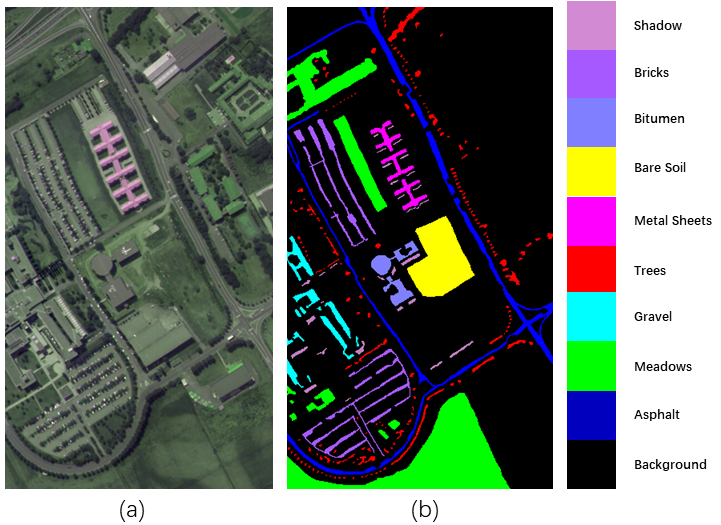} 
  \caption{University of Pavia(UP) dataset: (a) Pseudocolor image,(b) Ground-truth map}
  \label{fig:UP}
\end{figure}

\begin{figure}[!htbp] 
  \centering
  \small
  \includegraphics[width=1.05\linewidth]{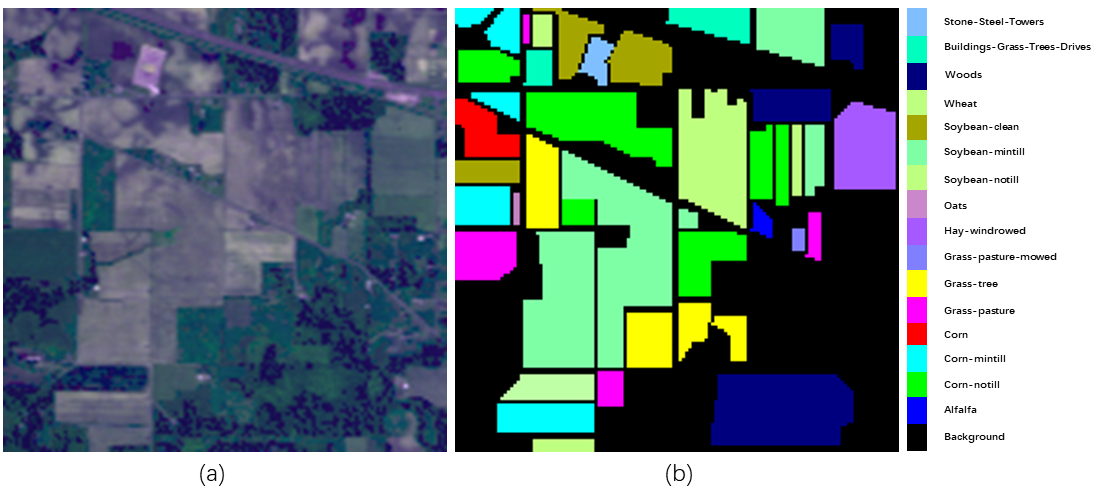} 
  \caption{Indian Pines(IP) dataset: (a) Pseudocolor image,(b) Ground-truth map}
  \label{fig:IP}
\end{figure}

\begin{figure}[!htbp] 
  \centering
  \small
  \includegraphics[width=0.75\linewidth]{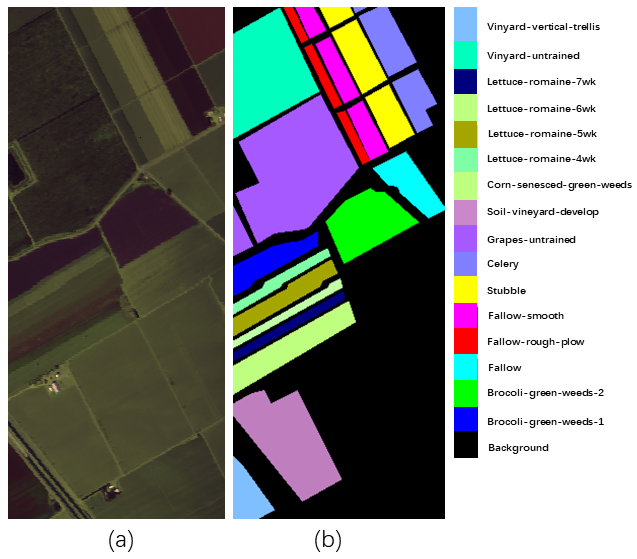} 
  \caption{Salinas(SA) dataset: (a) Pseudocolor image,(b) Ground-truth map}
  \label{fig:SA}
\end{figure}

\begin{figure}[!htbp] 
  \centering
  \small
  \includegraphics[width=0.67\linewidth]{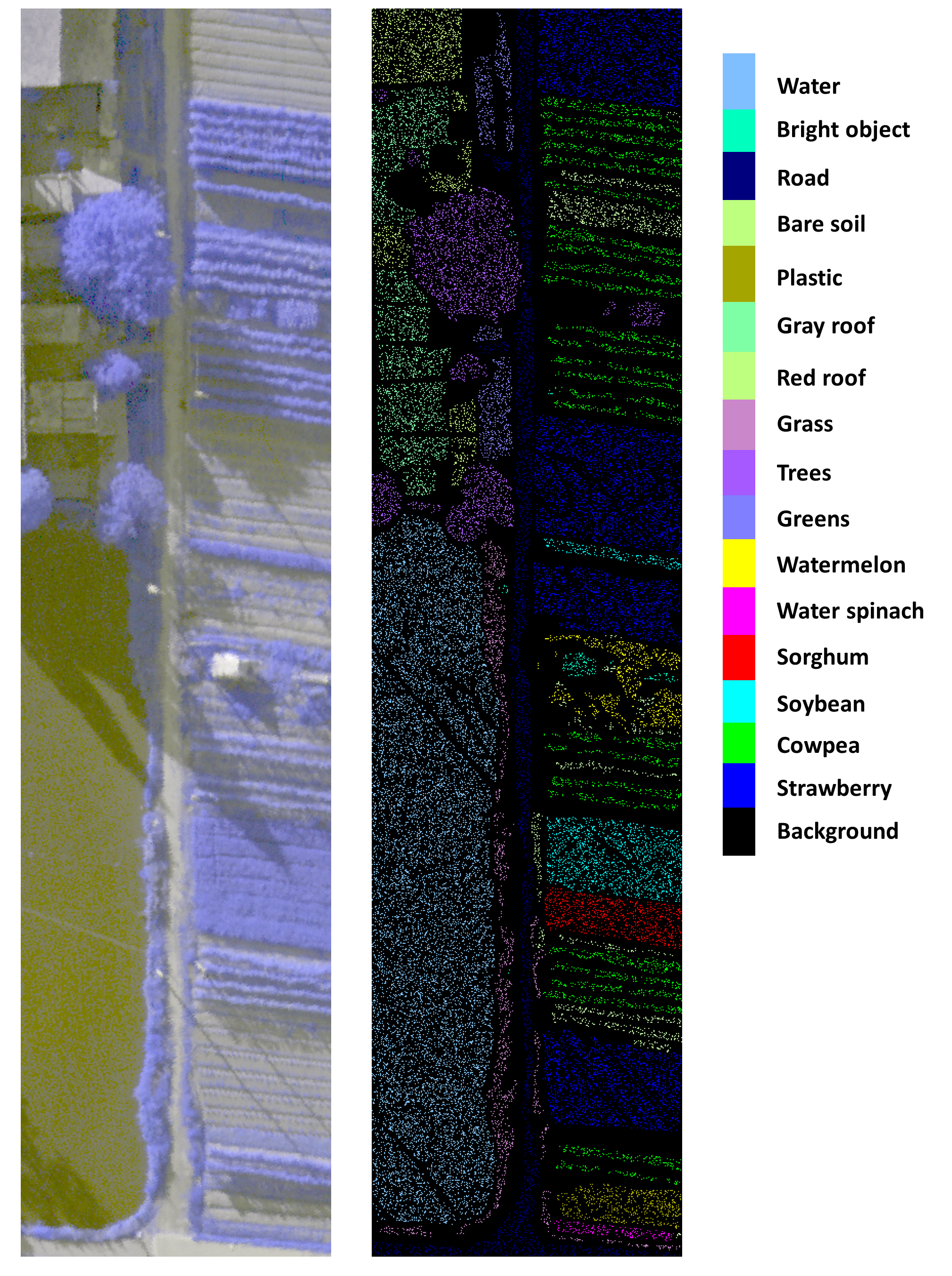} 
  \caption{WHU-Hi-HC(HC) dataset: (a) Pseudocolor image,(b) Ground-truth map}
  \label{fig:HC}
\end{figure}

Tables \ref{tab-UP}-\ref{tab-HC} present the land cover classes and sample counts for UP, IP, SA, and HC. Figures \ref{fig:UP}-\ref{fig:HC} show their pseudo-color images and ground truth maps.

\begin{table}[!htbp]
    \centering
    \small 
    \captionsetup{font=small} 
    \caption{Class Name and Corresponding Sample Number in the UP Dataset.}
    \label{tab-UP}
    \begin{tabularx}{\linewidth}{c >{\raggedright\arraybackslash}X c} 
    \toprule
    \textbf{Class} & \textbf{Name} & \textbf{Number of Samples} \\
    \midrule
    1 & Asphalt & 6631 \\
    2 & Meadows & 18649 \\
    3 & Gravel & 2099 \\
    4 & Trees & 3064 \\
    5 & Painted Metal & 1345 \\  
    6 & Bare Soil & 5029 \\
    7 & Bitumen & 1330 \\
    8 & Self-Blocking Bricks & 3682 \\ 
    9 & Shadows & 947 \\
    \midrule
    \textbf{Total} & & \textbf{42776} \\
    \bottomrule
    \end{tabularx}
\end{table}

\begin{table}[!htbp]
    \centering
    \small 
    \captionsetup{font=small} 
    \caption{Class Name and Corresponding Sample Number in the IP Dataset.} 
    \label{tab-IP} 
    \begin{tabularx}{\linewidth}{c >{\raggedright\arraybackslash}X c} 
    \toprule 
    \textbf{Class} & \textbf{Name} & \textbf{Number of Samples} \\ 
    \midrule 
    1 & Alfalfa & 46 \\
    2 & Corn-notill & 1428 \\
    3 & Corn-mintill & 830 \\
    4 & Corn & 237 \\
    5 & Grass-pasture & 483 \\
    6 & Grass-trees & 730 \\
    7 & Grass-pasture-mowed & 28 \\
    8 & Hay-windrowed & 478 \\
    9 & Oats & 20 \\
    10 & Soybean-notill & 972 \\
    11 & Soybean-mintill & 2455 \\
    12 & Soybean-clean & 593 \\
    13 & Wheat & 205 \\
    14 & Woods & 1265 \\
    15 & Buildings-Grass-Trees-Drives & 386 \\
    16 & Stone-Steel-Towers & 93 \\
    \midrule 
    \textbf{Total} & & \textbf{10249} \\ 
    \bottomrule 
    \end{tabularx}
\end{table}

\begin{table}[!htbp]
    \centering
    \small 
    \captionsetup{font=small} 
    \caption{Class Name and Corresponding Sample Number in the SA Dataset.} 
    \label{tab-SA} 
    \begin{tabularx}{\linewidth}{c >{\raggedright\arraybackslash}X c} 
    \toprule 
    \textbf{Class} & \textbf{Name} & \textbf{Number of Samples} \\ 
    \midrule 
    1 & Brocoli-green-weeds-1 & 2009 \\
    2 & Brocoli-green-weeds-2 & 3726 \\
    3 & Fallow & 1976 \\
    4 & Fallow-rough-plow & 1394 \\
    5 & Fallow-smooth & 2678 \\
    6 & Stubble & 3959 \\
    7 & Celery & 3579 \\
    8 & Grapes-untrained & 11271 \\
    9 & Soil-vineyard-develop & 6203 \\
    10 & Corn-senesced-green-weeds & 3278 \\
    11 & Lettuce-romaine-4wk & 1068 \\
    12 & Lettuce-romaine-5wk & 1927 \\
    13 & Lettuce-romaine-6wk & 916 \\
    14 & Lettuce-romaine-7wk & 1070 \\
    15 & Vinyard-untrained & 7268 \\
    16 & Vinyard-vertical-trellis & 1807 \\
    \midrule 
    \textbf{Total} & & \textbf{54129} \\ 
    \bottomrule 
    \end{tabularx}
\end{table}

\begin{table}[!htbp]
    \centering
    \small 
    \captionsetup{font=small} 
    \caption{Class Name and Corresponding Sample Number in the HC Dataset.} 
    \label{tab-HC} 
    \begin{tabularx}{\linewidth}{c >{\raggedright\arraybackslash}X c} 
    \toprule 
    \textbf{Class} & \textbf{Name} & \textbf{Number of Samples} \\ 
    \midrule 
    1 & Strawberry & 6710 \\
    2 & Cowpea & 3412 \\
    3 & Soybean & 1543 \\
    4 & Sorghum & 802 \\
    5 & Water spinach & 180 \\
    6 & Watermelon & 679 \\
    7 & Greens & 885 \\
    8 & Trees & 2696 \\
    9 & Grass & 1420 \\
    10 & Red roof & 1577 \\
    11 & Gray roof & 2536 \\
    12 & Plastic & 551 \\
    13 & Bare soil & 1367 \\
    14 & Road & 2784 \\
    15 & Bright object & 170 \\
    16 & Water & 11310 \\
    \midrule 
    \textbf{Total} & & \textbf{38622} \\ 
    \bottomrule 
    \end{tabularx}
\end{table}

\subsection{Experimental Settings}
To comprehensively evaluate model performance, we use three common evaluation metrics. These metrics are: Overall Accuracy (OA), Average Accuracy (AA), and the Kappa coefficient (Kappa). All reported experimental results are the average and standard deviation of 10 independent runs. This approach ensures the stability and reliability of our findings.

All experiments were conducted on a workstation equipped with an AMD Ryzen Threadripper 1950X 16-Core CPU, 128GB RAM, and an NVIDIA GeForce RTX 3090 GPU, using the PyTorch deep learning framework. 
 
Input HSI data is processed into image patches. Each patch is centered on a pixel and has dimensions of $33 \times 33 \times B$. Here, $B$ represents the number of bands for the corresponding dataset.

Stage 1: This stage involves heterogeneous data pre-training. The Mini-ImageNet dataset is used as the heterogeneous dataset $D_{s_1}$. The number of episodes is set to 1100. The batch size for the RM-SSL task is set to 128.

Stage 2: This stage involves homogeneous data pre-training. The Chikusei dataset is used as the homogeneous dataset $D_{s_2}$. The number of episodes is set based on the target dataset. For the UP, IP, SA, and HC datasets, the episode counts are 700, 500, 700, and 700, respectively. For the MR-SSL task, all samples from $D_{s_2}$ are used. The batch size is set to 1024, and the masking rate is set to 0.75. For the FSL task, samples are selected from 16 classes in the Chikusei dataset that have more than 400 samples. From each of these classes, 400 samples are provided to construct the support and query sets for FSL.

Stage 3: This stage is for few-shot fine-tuning on the target domain dataset $D_t$. The number of episodes is set to 1000. The batch size for the SSLCL task is $N_t \times K_0$. Here, $K_0$ represents the number of labeled samples per class in the target domain. In the experiments, $K_0$ ranges from 1 to 5. The Dropout layer ratio for SSLCL is set according to the target dataset. For the UP, IP, SA, and HC datasets, these ratios are 0.15, 0.28, 0.28, and 0.28, respectively. Testing and evaluation are performed periodically during training. For the SA and IP datasets, testing occurs every 50 episodes. For the UP and HC datasets, testing occurs every 20 episodes.

FSL Settings:
In all training stages involving FSL, the $C$-way $K$-shot episodic training method is adopted.
For a specific target domain dataset, the number of categories is $N_t$, specifically 9 on the UP dataset, 16 on IP, SA and HC. All FSL tasks corresponding to the target domain, including $N_{s_1}$ in stage 1, $N_{s_2}$ in stage 2 and $N_t$ in stage 3, are set to the number of target domain categories $N_t$. The number of support set samples $K$ is set to 1 in all training stages. The number of query samples $C$ contained in each category is set to 19.

Data Augmentation Settings:
To generate the $D_{l_1}$ data for target domain FSL, noise data augmentation is employed. This process first involves randomly generating a scaling factor $\alpha$ from a uniform distribution in the range $(0.9,1.1)$. Next, Gaussian noise with the same dimensions as the data is generated. This noise has a mean of 0 and a standard deviation of 1. The augmented data is then represented as $\alpha$ times the original data, plus the noise multiplied by a coefficient of $1/25$.

In the SSLCL task, the data augmentation function $\mathcal{T}(\cdot)$ is implemented as follows. With a probability of 0.5, random cropping is applied to the input HSI patch. The crop area ratio ranges from $(0.7,1.0)$, and the aspect ratio ranges from $(3/4,4/3)$. The cropped patch is then restored to its original size using bicubic interpolation. Alternatively, with a probability of 0.5, noise data augmentation is performed, using the same parameters as described above.
Furthermore, after random cropping is performed, the cropped result is subjected to additional transformations with a 0.5 probability. These transformations include random horizontal flipping, random vertical flipping, and random rotation by multiples of $90^\circ$.

Optimizer and Hyperparameters:
The Adam optimizer is used in all stages. The learning rate is fixed at 0.001. For combined losses in each stage and internal loss terms within SSLCL, direct summation is used.

\begin{table*}[!htbp]
    \centering
    \small
    \caption{Class-Specific Classification Accuracy (\%), OA (\%), AA (\%),
    and Kappa On The UP Dataset (Five Labeled Samples Per Class)\label{re-UP}}
    \begin{tabular*}{\textwidth}{@{\extracolsep{\fill}}lccccccccc} 
    \toprule
    \multicolumn{1}{c}{\multirow{2}{*}{\textbf{Class}}} & \multicolumn{1}{c}{\multirow{2}{*}{\textbf{SVM}}} & \multicolumn{1}{c}{\multirow{2}{*}{\textbf{SSRN}}} & \multicolumn{1}{c}{\multirow{2}{*}{\textbf{DCFSL}}} & \multicolumn{1}{c}{\textbf{Gia-}} & \multicolumn{1}{c}{\multirow{2}{*}{\textbf{HFSL}}} & \multicolumn{1}{c}{\multirow{2}{*}{\textbf{ADAFSL}}} & \multicolumn{1}{c}{{\textbf{FSCF-}}} & \multicolumn{1}{c}{\multirow{2}{*}{\textbf{FDFSL}}} & \multicolumn{1}{c}{\multirow{2}{*}{\textbf{S4L-FSC}}} \\
    & & & & \multicolumn{1}{c}{\textbf{CFSL}} & & &\multicolumn{1}{c}{\textbf{SSL}} & & \\ 
    \midrule
    \multicolumn{1}{c}{\hspace{0.2em} 1} & 57.18 & 76.77 & 80.77 & 81.10 & 73.08 & 81.79 & 83.05 & 86.68 & \textbf{90.13} \\ 
    \multicolumn{1}{c}{\hspace{0.2em} 2} & 60.00 & 72.58 & 85.79 & 89.11 & 90.98 & 82.66 & 91.60 & 82.73 & \textbf{92.49} \\
    \multicolumn{1}{c}{\hspace{0.2em} 3} & 41.07 & 72.49 & 61.54 & 64.49 & \textbf{88.33} & 69.72 & 79.35 & 62.47 & 82.50 \\
    \multicolumn{1}{c}{\hspace{0.2em} 4} & 82.50 & 82.21 & 93.54 & 91.66 & 93.37 & 92.38 & 86.81 & 91.35 & \textbf{95.47} \\
    \multicolumn{1}{c}{\hspace{0.2em} 5} & 98.26 & 99.87 & 98.48 & 99.63 & 99.87 & 97.99 & \textbf{99.98} & 99.11 & 99.81 \\
    \multicolumn{1}{c}{\hspace{0.2em} 6} & 53.70 & 77.67 & 76.77 & 73.20 & 88.65 & 77.30 & 87.91 & 71.24 & \textbf{93.07}  \\
    \multicolumn{1}{c}{\hspace{0.2em} 7} & 75.21 & 90.31 & 78.11 & 79.25 & 87.30 & 86.31 & 87.83 & 89.00 & \textbf{97.37}  \\
    \multicolumn{1}{c}{\hspace{0.2em} 8} & 69.27 & 71.74 & 67.44 & 73.84 & 94.81 & 74.76 & 94.65 & 88.04 & \textbf{97.77}  \\
    \multicolumn{1}{c}{\hspace{0.2em} 9} & \textbf{99.93} & 99.18 & 98.24 & 98.06 & 99.82 & 98.96 & 98.54 & 94.88 & 94.51  \\
    \midrule
    \multicolumn{1}{c}{\hspace{0.4em}\textbf{OA}} & 62.86 & 76.43 & 82.17 & 83.88 & 88.66 & 81.84 & 90.09 & 83.05 & \textbf{92.80}  \\
                       & $\pm$5.41  & $\pm$1.88 & $\pm$4.33 & $\pm$2.82 & $\pm$3.32 & $\pm$4.47 & $\pm$3.81 & $\pm$2.52 & \textbf{$\pm$2.35} \\ 
    \multicolumn{1}{c}{\hspace{0.4em}\textbf{AA}} & 70.79 & 82.54 & 82.30 & 83.37 & 90.69 & 84.04 & 90.95 & 85.06 & \textbf{93.68}  \\
                       & $\pm$2.57 & $\pm$4.53 & $\pm$2.58 & $\pm$2.32 & $\pm$2.14 & $\pm$2.23 & $\pm$3.33 & $\pm$2.30 & \textbf{$\pm$2.65} \\ 
    \multicolumn{1}{c}{\textbf{Kappa}} & 53.77 & 69.94 & 76.92 & 78.95 & 85.24 & 76.58 & 87.07 & 78.04 & \textbf{90.58} \\
                       & $\pm$5.30 & $\pm$2.29 & $\pm$5.12 & $\pm$3.44 & $\pm$4.08 & $\pm$5.18 & $\pm$4.84 & $\pm$2.80 & \textbf{$\pm$3.01}  \\ 
    \bottomrule
    \end{tabular*}
\end{table*}

\begin{table*}[!htbp]
    \centering
    \small
    \caption{Class-Specific Classification Accuracy (\%), OA (\%), AA (\%),
    and Kappa On The IP Dataset (Five Labeled Samples Per Class)\label{re-IP}}
    \begin{tabular*}{\textwidth}{@{\extracolsep{\fill}}c ccc c ccc c c} 
    \toprule
    \multirow{2}{3em}{\centering \textbf{Class}} & \multicolumn{1}{c}{\multirow{2}{*}{\textbf{SVM}}} & \multicolumn{1}{c}{\multirow{2}{*}{\textbf{SSRN}}} & \multicolumn{1}{c}{\multirow{2}{*}{\textbf{DCFSL}}} & \multicolumn{1}{c}{\textbf{Gia-}} & \multicolumn{1}{c}{\multirow{2}{*}{\textbf{HFSL}}} & \multicolumn{1}{c}{\multirow{2}{*}{\textbf{ADAFSL}}} & \multicolumn{1}{c}{{\textbf{FSCF-}}} & \multicolumn{1}{c}{\multirow{2}{*}{\textbf{FDFSL}}} & \multicolumn{1}{c}{\multirow{2}{*}{\textbf{S4L-FSC}}} \\
    & & & & \multicolumn{1}{c}{\textbf{CFSL}} & & & \multicolumn{1}{c}{\textbf{SSL}}  & &  \\ 
    \midrule
    1 & 70.73 & 92.44 & 92.44 & 92.68 & 99.02 & 98.05 & 99.51 & 96.34 & \textbf{99.76} \\
    2 & 37.59 & 53.16 & 41.15 & 46.90 & 56.58 & 51.99 & 54.96 & 52.04 & \textbf{61.28} \\
    3 & 43.93 & 40.78 & 48.95 & 46.50 & 64.25 & 65.24 & 66.23 & 66.04 & \textbf{69.62} \\
    4 & 49.83 & 63.41 & 72.63 & 72.24 & 91.34 & 81.29 & 91.68 & 84.27 & \textbf{97.54} \\
    5 & 62.28 & 70.79 & 73.77 & 74.08 & 71.26 & \textbf{78.41} & 72.74 & 73.62 & 75.84 \\
    6 & 67.01 & 84.65 & 85.10 & 83.14 & 78.99 & 86.91 & 80.08 & \textbf{88.68} & 79.90 \\
    7 & 80.43 & \textbf{100} & 99.13 & 99.57 & 99.57 & \textbf{100} & \textbf{100} & \textbf{100} & \textbf{100}  \\
    8 & 58.54 & 93.55 & 83.93 & 89.92 & 97.72 & 88.63 & 99.01 & 85.24 & \textbf{99.89}  \\
    9 & 90.00 & 98.67 & 98.67 & \textbf{100} & \textbf{100} & \textbf{100} & \textbf{100} & \textbf{100} & \textbf{100}  \\
    10 & 41.32 & 49.72 & 64.41 & 55.73 & 61.91 & \textbf{72.11} & 64.71 & 64.59 & 65.82 \\
    11 & 39.16 & 45.02 & 60.73 & 60.35 & 72.92 & 59.15 & 74.08 & 63.27 & \textbf{75.85} \\
    12 & 26.55 & 52.81 & 47.18 & 46.65 & \textbf{72.07} & 56.05 & 70.78 & 57.38 & 63.18 \\
    13 & 89.90 & 87.90 & 97.75 & 99.55 & 98.45 & 98.05 & \textbf{99.70} & 98.75 & 97.95  \\
    14 & 66.31 & \textbf{92.47} & 85.01 & 81.17 & 91.09 & 85.06 & 90.47 & 87.15 & 91.94  \\
    15 & 28.58 & 78.45 & 74.38 & 74.86 & 92.20 & 78.40 & 91.84 & 82.57 & \textbf{95.54}  \\
    16 & 88.86 & \textbf{99.43} & \textbf{99.43} & 98.30 & 92.95 & 98.41 & 95.11 & 98.07 & 96.59  \\
    \midrule
    \multirow{2}{3em}{\centering \textbf{OA}} & 47.71 & 62.27 & 65.15 & 64.53 & 74.62 & 69.80 & 75.19 & 70.46 & \textbf{77.05}   \\
                       & $\pm$2.19  & $\pm$4.11 & $\pm$3.10 & $\pm$3.56 & $\pm$3.06 & $\pm$4.01 & $\pm$2.92 & $\pm$1.91 & \textbf{$\pm$3.12} \\
    \multirow{2}{3em}{\centering \textbf{AA}} & 58.81 & 75.20 & 76.54 & 76.35 & 83.77 & 81.11 & 84.43 & 81.13 & \textbf{85.67}  \\
                       & $\pm$1.50 & $\pm$4.36 & $\pm$1.43 & $\pm$1.66 & $\pm$2.59 & $\pm$1.73 & $\pm$2.25 & $\pm$1.42 & \textbf{$\pm$1.50}  \\
    \multirow{2}{3em}{\centering \textbf{Kappa}} & 41.53 & 57.55 & 60.72 & 59.98 & 71.31 & 66.08 & 71.95 & 66.81 & \textbf{74.17} \\
                       & $\pm$2.10 & $\pm$4.56 & $\pm$3.14 & $\pm$3.83 & $\pm$3.44 & $\pm$4.22 & $\pm$3.21 & $\pm$2.13 & \textbf{$\pm$3.44}  \\
    \bottomrule
    \end{tabular*}
\end{table*}

\subsection{Experiment Results}
We compare our proposed S4L-FSC method with several related HSI classification methods, including SVM \cite{1323134}, SSRN \cite{zhong2017spectral}, DCFSL \cite{dcfsl}, GIA-CFSL \cite{giacfsl}, ADAFSL \cite{adafsl}, HFSL \cite{hfsl}, FSCF-SSL \cite{fscf-ssl}, and FDFSL \cite{fdfsl}. For evaluation, we randomly select 1 to 5 labeled samples per class from the four datasets as training data, and the rest are used for testing. To ensure the stability of the experimental results, we take the average of 10 runs, and the results are listed in Tables \ref{re-UP}–\ref{re-HC}.

S4L-FSC outperforms all methods on various datasets. On the UP dataset, S4L-FSC outperforms the second-ranked FSCF-SSL by 2.71\%, 2.73\%, and 3.51\% in OA, AA, and Kappa, respectively, and achieves the highest per-class accuracy in all 9 categories. On the IP dataset, despite the low-resolution and high-variability issues of the IP dataset, S4L-FSC outperforms FSCF-SSL by 1.86\%, 1.24\%, and 2.22\% in OA, AA, and Kappa, respectively, and performs well in 9 categories. On the SA dataset, S4L-FSC outperforms FSCF-SSL by 2.21\% in OA, 0.49\% in AA, and 2.45\% in Kappa, respectively, and leads in 8 categories. The most significant improvement is on the HC dataset, where its OA, AA, and Kappa are 5.68\%, 7.44\%, and 6.58\% higher than FSCF-SSL, respectively, and achieves the highest accuracy in 8 categories.

\begin{figure*}[!htbp]
    \centering
    \small
    \includegraphics[width=0.8\textwidth]{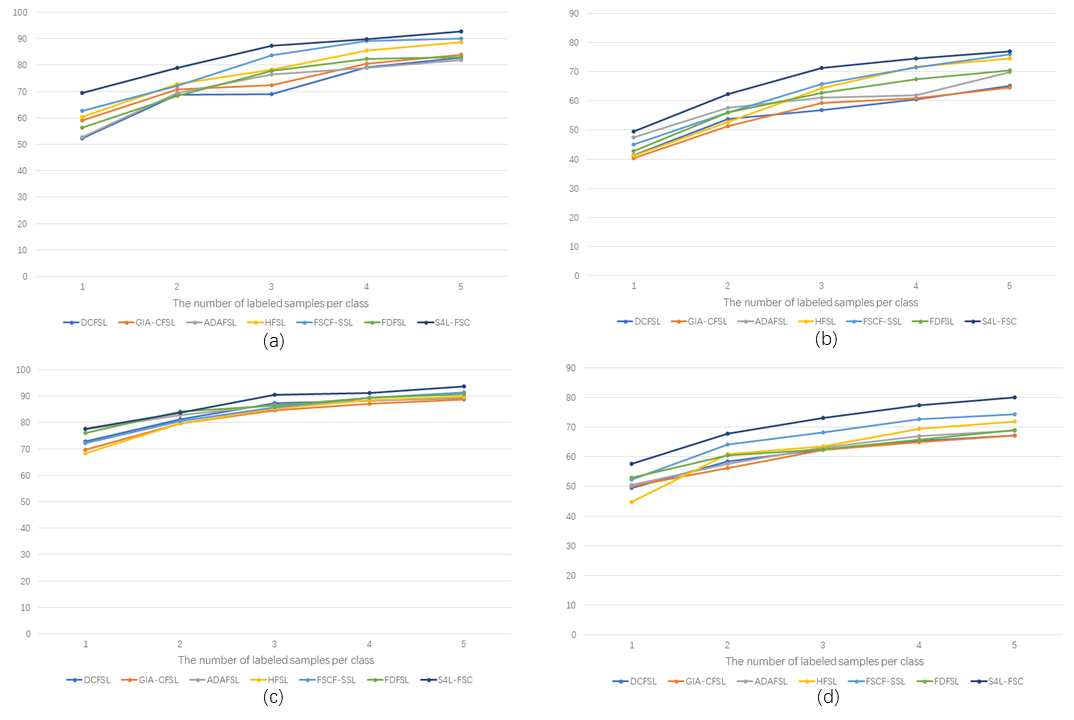}
    \caption{Classification results of all comparison methods under different labeled samples on (a) UP, (b) IP, (c) SA, and (d) HC datasets}
\label{labeled_4}
\end{figure*}

Extended experiments with 1–5 labeled samples, as shown in Figure~\ref{labeled_4}, confirmed S4L-FSC’s robustness, achieving optimal performance in most cases, except on SA with 2 samples, where FDFSL slightly led in OA. Notably, on UP and HC with 1–2 samples, S4L-FSC significantly outperformed competitors like FSCF-SSL and HFSL.

\begin{figure*}[!t]
\centering
\small
\includegraphics[width=0.45\textwidth]{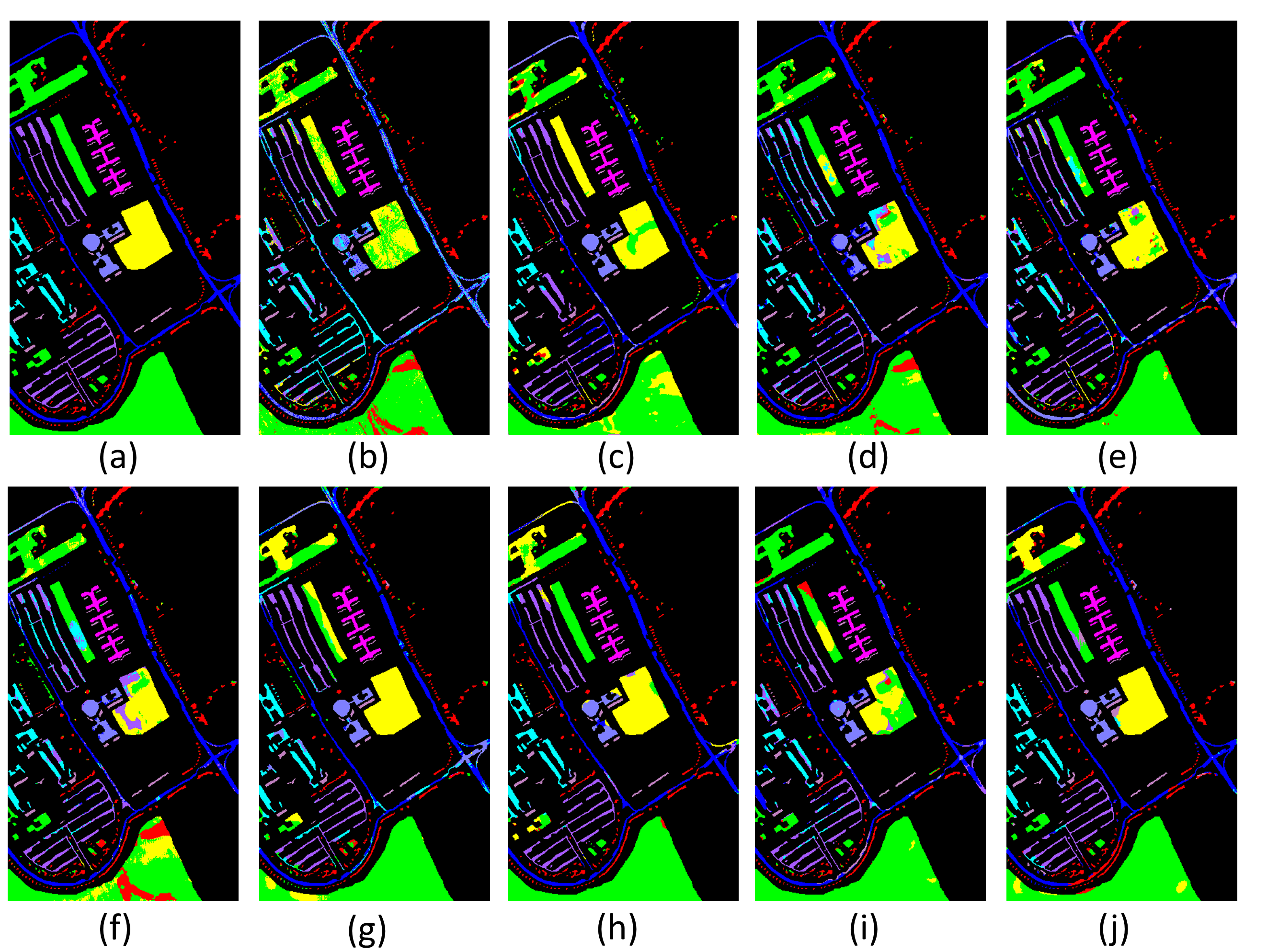}
\caption{Classification maps obtained using different methods on the UP dataset. (a) Ground truth. (b) SVM. (c) SSRN. (d) DCFSL. (e) GIA-CFSL. (f) ADAFSL. (g) HFSL. (h) FSCF-SSL. (i) FDFSL. (j) S4L-FSC.}
\label{re_UP10}
\end{figure*}

\begin{figure*}[!t]
\centering
\small
\includegraphics[width=0.6\textwidth]{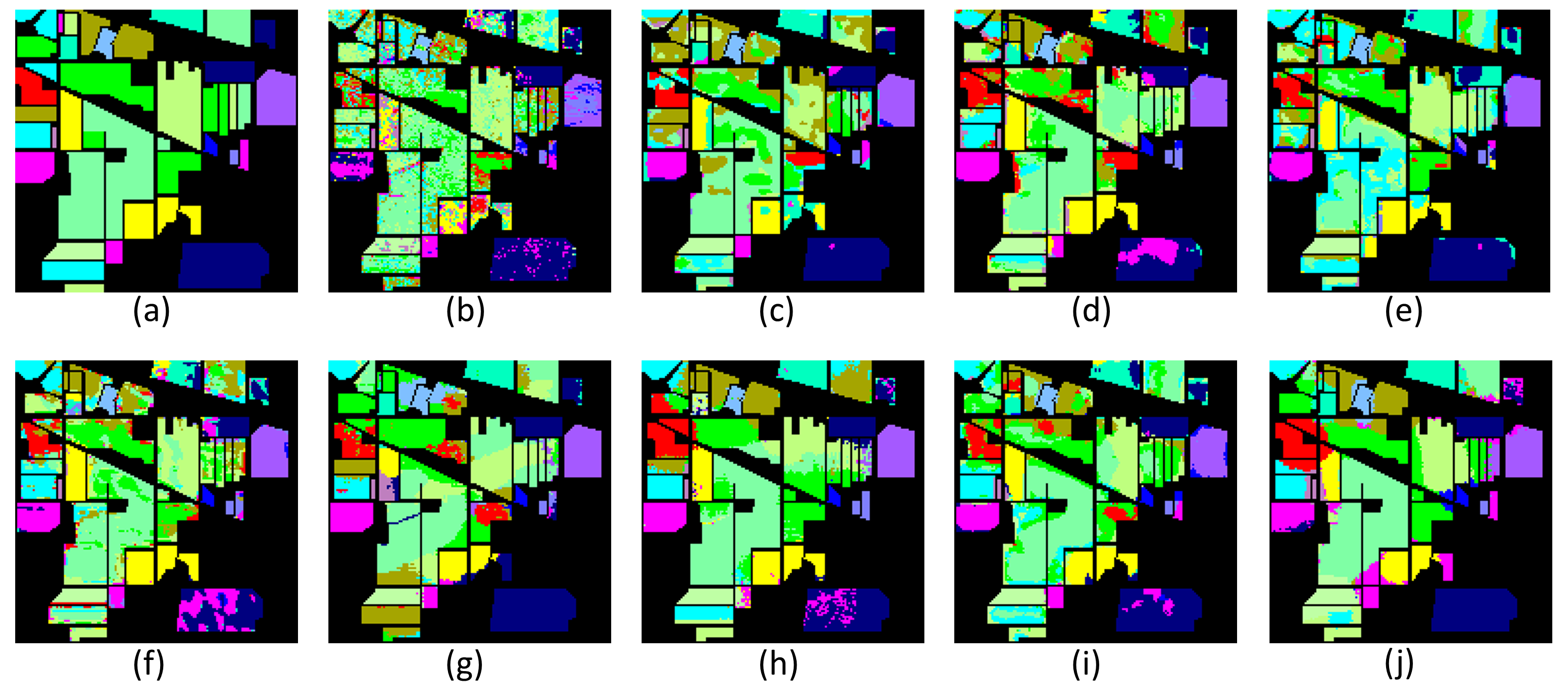}
\caption{Classification maps obtained using different methods on the IP dataset. (a) Ground truth. (b) SVM. (c) SSRN. (d) DCFSL. (e) GIA-CFSL. (f) ADAFSL. (g) HFSL. (h) FSCF-SSL. (i) FDFSL. (j) S4L-FSC.}
\label{re_IP10}
\end{figure*}

\begin{figure*}[!t]
\centering
\small
\includegraphics[width=0.4\textwidth]{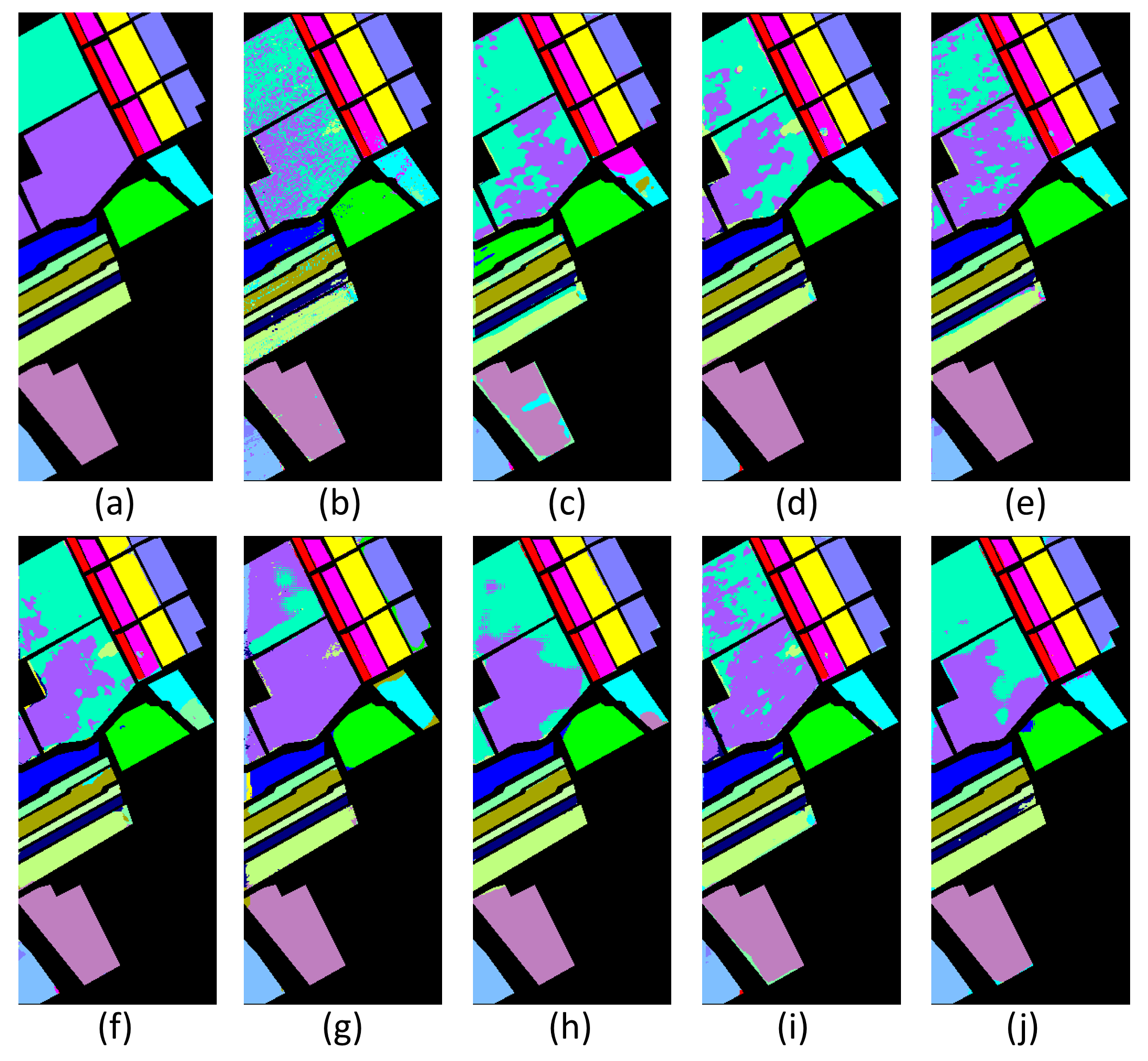}
\caption{Classification maps obtained using different methods on the SA dataset. (a) Ground truth. (b) SVM. (c) SSRN. (d) DCFSL. (e) GIA-CFSL. (f) ADAFSL. (g) HFSL. (h) FSCF-SSL. (i) FDFSL. (j) S4L-FSC.}
\label{re_SA10}
\end{figure*}

\begin{figure*}[!t]
\centering
\small
\includegraphics[width=0.4\textwidth]{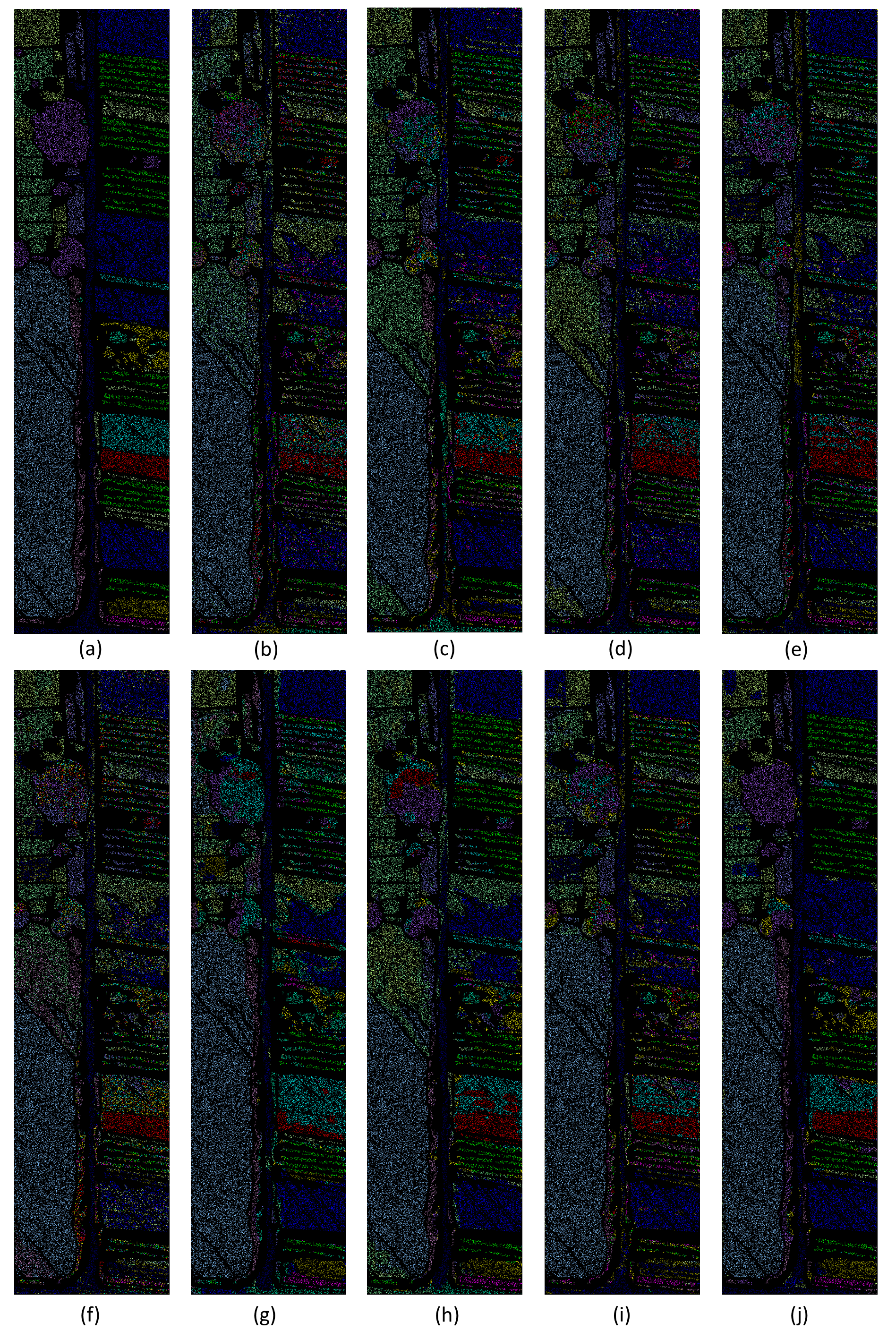}
\caption{Classification maps obtained using different methods on the HC dataset. (a) Ground truth. (b) SVM. (c) SSRN. (d) DCFSL. (e) GIA-CFSL. (f) ADAFSL. (g) HFSL. (h) FSCF-SSL. (i) FDFSL. (j) S4L-FSC.}
\label{re_HC10}
\end{figure*}

Classification maps,  as shown in Figures \ref{re_UP10}-\ref{re_HC10}, with 5 labeled samples per class visually reinforce these results. S4L-FSC’s outputs closely match ground truth, with fewer misclassified pixels and sharper class boundaries compared to methods like FSCF-SSL, particularly for classes like Asphalt in UP and Grapes-untrained in SA.

In summary, S4L-FSC excels across all datasets, demonstrating the effectiveness of its self-supervised pre-training with heterogeneous and homogeneous data, followed by target-domain fine-tuning, for few-shot HSI classification.

\begin{table*}[!t]
    \centering
    \small
    \caption{Class-Specific Classification Accuracy (\%), OA (\%), AA (\%),
    and Kappa On The SA Dataset (Five Labeled Samples Per Class)\label{re-SA}}
    \begin{tabular*}{\textwidth}{@{\extracolsep{\fill}}c c c c c c c c c c} 
    \toprule
    \multirow{2}{*}{\centering \textbf{Class}} & \multicolumn{1}{c}{\multirow{2}{*}{\textbf{SVM}}} & \multicolumn{1}{c}{\multirow{2}{*}{\textbf{SSRN}}} & \multicolumn{1}{c}{\multirow{2}{*}{\textbf{DCFSL}}} & \multicolumn{1}{c}{\textbf{Gia-}} & \multicolumn{1}{c}{\multirow{2}{*}{\textbf{HFSL}}}  & \multicolumn{1}{c}{\multirow{2}{*}{\textbf{ADAFSL}}} & \multicolumn{1}{c}{\textbf{FSCF-}} & \multicolumn{1}{c}{\multirow{2}{*}{\textbf{FDFSL}}} & \multicolumn{1}{c}{\multirow{2}{*}{\textbf{S4L-FSC}}} \\
    & & & & \multicolumn{1}{c}{\textbf{CFSL}} & & & \multicolumn{1}{c}{\textbf{SSL}} & &  \\ 
    \midrule
    \multicolumn{1}{c}{1} & 98.15 & 62.22 & 99.36 & 99.56 & 97.81 & 99.22 & \textbf{99.96} & 98.96 & 99.56 \\
    \multicolumn{1}{c}{2} & 95.03 & 96.50 & 99.55 & 98.52 & 93.91 & 99.51 & 99.16 & 98.99 & \textbf{99.98} \\
    \multicolumn{1}{c}{3} & 81.75 & 40.76 & 95.00 & 88.31 & 98.18 & 93.06 & 95.66 & 90.50 & \textbf{99.03} \\
    \multicolumn{1}{c}{4} & 97.84 & 98.32 & 99.12 & 98.71 & 99.58 & 99.11 & 99.68 & 99.34 & \textbf{99.87} \\
    \multicolumn{1}{c}{5} & 95.11 & 95.31 & 92.51 & 89.13 & 95.82 & 93.28 & \textbf{96.98} & 91.96 & 96.27 \\
    \multicolumn{1}{c}{6} & 99.01 & 99.27 & 99.09 & 98.11 & 99.64 & 97.73 & \textbf{99.47} & 99.07 & 99.03 \\ 
    \multicolumn{1}{c}{7} & 99.12 & 93.42 & 98.60 & 99.08 & 99.26 & 97.83 & 98.93 & 99.31 & \textbf{99.44}  \\
    \multicolumn{1}{c}{8} & 53.47 & 73.29 & 75.05 & 76.72 & 76.48 & 75.70 & 78.64 & 79.24 & \textbf{86.09}  \\ 
    \multicolumn{1}{c}{9} & 98.17 & 85.77 & 99.41 & 98.42 & 99.31 & 99.62 & 99.23 & 98.41 & \textbf{99.77}  \\
    \multicolumn{1}{c}{10} & 77.11 & 66.99 & 87.15 & 82.21 & 91.12 & 85.20 & \textbf{95.06} & 83.77 & 93.53 \\
    \multicolumn{1}{c}{11} & 88.33 & 69.30 & 98.13 & 95.27 & 98.97 & 98.27 & \textbf{99.60} & 96.54 & 97.32 \\
    \multicolumn{1}{c}{12} & 93.29 & 95.32 & \textbf{99.68} & 98.39 & 96.07 & 99.47 & 99.48 & 99.37 & 97.55 \\
    \multicolumn{1}{c}{13} & 95.64 & 98.24 & 99.22 & 97.78 & 99.03 & 99.54 & \textbf{99.56} & 98.73 & 98.24  \\
    \multicolumn{1}{c}{14} & 90.64 & 90.63 & 98.80 & 98.02 & 93.41 & 97.99 & 97.59 & \textbf{99.20} & 96.58  \\
    \multicolumn{1}{c}{15} & 52.28 & 71.32 & 74.09 & 75.51 & 75.57 & 76.87 & 76.81 & 81.46 & \textbf{82.30}  \\
    \multicolumn{1}{c}{16} & 83.57 & 93.05 & 90.84 & 89.94 & 91.64 & 92.10 & \textbf{98.92} & 92.61 & 98.00  \\
    \midrule
    \multirow{2}{*}{\centering \textbf{OA}} & 79.52 & 81.16 & 89.29 & 88.72 & 89.75 & 89.51 & 91.44 & 90.66 & \textbf{93.65}   \\
                       & $\pm$2.40  & $\pm$3.32 & $\pm$2.07 & $\pm$1.49 & $\pm$1.10 & $\pm$2.50 & $\pm$1.82 & $\pm$1.44 & \textbf{$\pm$0.89} \\
    \multirow{2}{*}{\centering \textbf{AA}} & 87.41 & 83.11 & 94.10 & 92.73 & 94.11 & 94.02 & 95.92 & 94.22 & \textbf{96.41}  \\
                       & $\pm$1.60 & $\pm$3.32 & $\pm$0.83 & $\pm$1.63 & $\pm$0.96 & $\pm$1.26 & $\pm$1.02 & $\pm$1.00 & \textbf{$\pm$0.63}  \\
    \multirow{2}{*}{\centering \textbf{Kappa}} & 77.29 & 79.07 & 88.10 & 87.47 & 88.63 & 88.34 & 90.49 & 89.63 & \textbf{92.94} \\
                       & $\pm$2.64 & $\pm$3.68 & $\pm$2.26 & $\pm$1.63 & $\pm$1.22 & $\pm$2.75 & $\pm$2.03 & $\pm$1.60 & \textbf{$\pm$0.99}  \\
    \bottomrule
    \end{tabular*}
\end{table*}

\begin{table*}[!t]
    \centering
    \small
    \caption{Class-Specific Classification Accuracy (\%), OA (\%), AA (\%),
    and Kappa On The HC Dataset (Five Labeled Samples Per Class)\label{re-HC}}
    \begin{tabular*}{\textwidth}{@{\extracolsep{\fill}}lccccccccc}
    \toprule
    \multirow{2}{*}{\textbf{Class}} & \multicolumn{1}{c}{\multirow{2}{*}{\textbf{SVM}}} & \multicolumn{1}{c}{\multirow{2}{*}{\textbf{SSRN}}} & \multicolumn{1}{c}{\multirow{2}{*}{\textbf{DCFSL}}} & \multicolumn{1}{c}{\textbf{Gia-}} & \multicolumn{1}{c}{\multirow{2}{*}{\textbf{HFSL}}}  & \multicolumn{1}{c}{\multirow{2}{*}{\textbf{ADAFSL}}} & \multicolumn{1}{c}{\textbf{FSCF-}} & \multicolumn{1}{c}{\multirow{2}{*}{\textbf{FDFSL}}} & \multicolumn{1}{c}{\multirow{2}{*}{\textbf{S4L-FSC}}} \\
    & & & & \multicolumn{1}{c}{\textbf{CFSL}} &  & & \multicolumn{1}{c}{\textbf{SSL}} & & \\ 
    \midrule
    \hspace{0.2em} 1 & 58.42 & 74.56 & 75.60 & 70.20 & 68.38 & 76.07 & 74.91 & 75.54 & \textbf{80.08} \\
    \hspace{0.2em} 2 & 26.20 & 42.40 & 41.47 & 48.55 & 70.41 & 35.72 & 82.45 & 45.76 & \textbf{83.33} \\
    \hspace{0.2em} 3 & 51.92 & 65.40 & 56.22 & \textbf{88.31} & 61.89 & 74.64 & 69.34 & 62.37 & 81.21 \\
    \hspace{0.2em} 4 & 73.58 & 90.83 & 90.49 & 91.08 & 86.32 & \textbf{94.02} & 88.43 & 85.67 & 93.02 \\
    \hspace{0.2em} 5 & 48.11 & 74.63 & 90.74 & 81.49 & 90.17 & 83.89 & \textbf{99.83} & 91.77 & 98.46 \\
    \hspace{0.2em} 6 & 19.39 & 33.04 & 27.15 & 30.62 & 45.80 & 36.60 & 45.77 & 36.84 & \textbf{58.18} \\ 
    \hspace{0.2em} 7 & 78.06 & 70.14 & 80.28 & 76.10 & 70.25 & 77.63 & 65.95 & 70.45 & \textbf{80.32}  \\
    \hspace{0.2em} 8 & 29.99 & 34.92 & 39.02 & 39.02 & 43.93 & 35.20 & 53.66 & 44.25 & \textbf{65.66}  \\ 
    \hspace{0.2em} 9 & 20.49 & 40.62 & 34.46 & 33.25 & 54.12 & 46.97 & \textbf{56.13} & 35.92 & 55.77  \\
    \hspace{0.2em} 10 & 60.48 & 64.25 & 69.38 & 70.39 & 53.38 & \textbf{79.88} & 48.37 & 75.22 & 79.71 \\
    \hspace{0.2em} 11 & 81.87 & \textbf{91.34} & 73.62 & 66.03 & 58.23 & 72.45 & 60.15 & 55.38 & 75.26 \\
    \hspace{0.2em} 12 & 33.63 & 38.61 & 48.15 & 48.11 & 77.69 & 51.41 & 82.56 & 53.44 & \textbf{86.54} \\
    \hspace{0.2em} 13 & 32.62 & 25.44 & 38.01 & 43.41 & 39.27 & 44.96 & 44.65 & 34.36 & \textbf{52.03}  \\
    \hspace{0.2em} 14 & 48.61 & 52.86 & 61.70 & 64.79 & 73.44 & 60.58 & \textbf{76.47} & 59.91 & 76.24  \\
    \hspace{0.2em} 15 & 58.91 & 70.73 & 71.70 & 70.73 & 79.03 & 51.33 & 86.55 & 68.24 & \textbf{87.88}  \\
    \hspace{0.2em} 16 & 72.15 & 70.12 & 86.58 & 87.50 & \textbf{93.98} & 87.54 & 89.34 & 92.84 & 89.99  \\
    \midrule
    \multirow{2}{*}{\hspace{0.4em}\textbf{OA}} & 55.65 & 62.40 & 67.28 & 67.28 & 71.93 & 68.89 & 74.24 & 69.05 & \textbf{79.92}   \\
                       & $\pm$2.74  & $\pm$1.70 & $\pm$4.02 & $\pm$3.81 & $\pm$4.23 & $\pm$3.80 & $\pm$5.77 & $\pm$2.89 & \textbf{$\pm$4.69} \\
    \multirow{2}{*}{\hspace{0.4em}\textbf{AA}} & 49.65 & 58.74 & 61.22 & 61.09 & 66.64 & 63.06 & 70.29 & 61.75 & \textbf{77.73}  \\
                       & $\pm$1.94 & $\pm$3.41 & $\pm$2.30 & $\pm$2.95 & $\pm$5.19 & $\pm$2.83 & $\pm$3.87 & $\pm$1.79 & \textbf{$\pm$3.47}  \\
    \multirow{2}{*}{\textbf{Kappa}} & 50.01 & 57.34 & 62.45 & 62.49 & 67.64 & 64.36 & 70.28 & 64.23 & \textbf{76.86} \\
                       & $\pm$2.87 & $\pm$1.92 & $\pm$4.35 & $\pm$4.06 & $\pm$4.77 & $\pm$3.98 & $\pm$6.22 & $\pm$3.07 & \textbf{$\pm$5.25}  \\
    \bottomrule
    \end{tabular*}
\end{table*}

\subsection{Ablation Experiment}
To verify the effectiveness of each key component in the S4L-FSC model, we conducted a series of ablation experiments.

First, we evaluated the contribution of S4L-FSC's core modules, with results presented in Table~\ref{tab:ablation_modules1_centered}. All tested variations resulted in performance degradation compared to the full S4L-FSC model. These included variant $v_0$, where the RM-SSL module was replaced with SSLGT from FSCF-SSL \cite{fscf-ssl}; variant $v_2$, which lacked homogeneous data FSL; variant $v_3$, which was without MR-SSL; and variant $v_4$, which excluded target domain SSLCL. The most significant performance drop occurred with variant $v_1$, which simultaneously removed both homogeneous data pre-training modules, namely homogeneous data FSL and MR-SSL. This highlights the crucial role of these complementary tasks during homogeneous pre-training for learning rich spectral priors and general-purpose features from large-scale data, establishing a strong foundation for target domain classification. Furthermore, the notable performance decrease with variant $v_4$, which removed the target domain SSLCL module, underscores its importance in effectively utilizing limited labeled samples in the target domain. Overall, each module in S4L-FSC positively contributes to its final performance.

Second, we specifically analyzed our proposed SSLCL method against two alternatives: variant $v_a$, which utilized the SSLCL approach from FSCF-SSL, and variant $v_b$, a baseline without any SSLCL module. Results are detailed in Table~\ref{tab:sslcl_comparison_v2_centered}. Our SSLCL strategy consistently outperformed both variant $v_a$ and $v_b$ across all test datasets, achieving the highest classification accuracy. The superiority of our SSLCL method stems from its active construction of semantically richer and more distinct sample views via data augmentation techniques, in contrast to methods relying solely on internal network dropout. This strategy increases input feature diversity, thereby promoting the learning of more robust and discriminative features, which ultimately leads to superior classification performance.

In summary, the ablation studies fully verify the effectiveness and necessity of each module proposed in the S4L-FSC model. These modules are crucial for improving the performance of few-shot hyperspectral image classification.

\begin{table*}[htbp]
    \centering
    \small 
    \setlength{\tabcolsep}{0.5pt} 
    \caption{The ablation experiment of S4L-FSC module}
    \label{tab:ablation_modules1_centered} 
    \begin{tabular*}{\textwidth}{@{\extracolsep{\fill}}ccccccccc} 
    \toprule
    \textbf{Method} & \textbf{RM-SSL} & \textbf{Homogeneous Data FSL} & \textbf{MR-SSL} & \textbf{SSLCL} & \textbf{UP} & \textbf{IP} & \textbf{SA} & \textbf{HC} \\
    \midrule 
    S4L-FSC        & \cmark    & \cmark    & \cmark    & \cmark    & \textbf{92.80} & \textbf{77.05} & \textbf{93.65} & \textbf{79.92} \\
    \(v_0\)   & \xmark    & \cmark    & \cmark    & \cmark    & 92.03          & 75.84          & 93.22          & 79.51          \\
    \(v_1\)   & \cmark    & \xmark    & \xmark    & \cmark    & 90.88          & 76.29          & 93.26          & 78.89          \\
    \(v_2\)   & \cmark    & \xmark    & \cmark    & \cmark    & 92.30          & 76.84          & 93.15          & 79.67          \\
    \(v_3\)   & \cmark    & \cmark    & \xmark    & \cmark    & 92.71          & 76.88          & 93.39          & 79.84          \\
    \(v_4\)   & \cmark    & \cmark    & \cmark    & \xmark    & 89.05          & 72.75          & 90.74          & 74.77          \\
    \bottomrule
    \end{tabular*}
\end{table*}

\begin{table*}[htbp]
    \small
    \centering
    \caption{Comparison of different SSLCL methods}
    \label{tab:sslcl_comparison_v2_centered} 
    \setlength{\tabcolsep}{5pt}
    \renewcommand{\arraystretch}{1.2}
    \setlength{\heavyrulewidth}{1pt}
    \setlength{\lightrulewidth}{0.7pt}

    \begin{tabular*}{\textwidth}{@{\extracolsep{\fill}}ccccccc} 
    \toprule

    \textbf{Method} & \textbf{SSLCL in S4L-FSC} & \textbf{SSLCL in FSCF-SSL} & \textbf{UP} & \textbf{IP} & \textbf{SA} & \textbf{HC} \\
    \midrule

    S4L-FSC & \cmark    & \xmark    & \textbf{92.80} & \textbf{77.05} & \textbf{93.65} & \textbf{79.92} \\

    \(v_a\)     & \xmark    & \cmark    & 89.40          & 74.69          & 91.66          & 74.36          \\

    \(v_b\)       & \xmark    & \xmark    & 89.05          & 72.75          & 90.74          & 74.77          \\
    \bottomrule
    \end{tabular*}
\end{table*}

\begin{table*}[!htbp]
\centering
\small
\caption{Computational Complexity Comparison: Training Time (s), Testing Time (s), and Number of Parameters on Four Datasets}
\label{tab:complexity}
\renewcommand{\arraystretch}{0.85} 
\begin{tabularx}{\textwidth}{c >{\centering\arraybackslash}p{1.4cm} *{7}{>{\centering\arraybackslash}X}} 

\toprule
\multirow{2}{*}{\textbf{Dataset}} & \multirow{2}{1.5cm}{\centering\textbf{Metric}} 
& \multirow{2}{*}{\textbf{DCFSL}} & \multirow{2}{*}{\textbf{\shortstack[c]{GIA- \\ CFSL}}} 
& \multirow{2}{*}{\textbf{ADAFSL}} & \multirow{2}{*}{\textbf{HFSL}} 
& \multirow{2}{*}{\textbf{\shortstack[c]{FSCF- \\ SSL}}} & \multirow{2}{*}{\textbf{FDFSL}} 
& \multirow{2}{*}{\textbf{S4L-FSC}} \\[1.8ex] 
\midrule

\multirow{5}{*}{UP} 
& Training Time & \multirow{2}{*}{1118.7} & \multirow{2}{*}{2647.0} & \multirow{2}{*}{1626.6} 
& \multirow{2}{*}{360.3} & \multirow{2}{*}{1334.7} & \multirow{2}{*}{707.4} & \multirow{2}{*}{1426.4} \\
& Testing Time  & \multirow{2}{*}{4.89}  & \multirow{2}{*}{4.64}  & \multirow{2}{*}{5.70} 
& \multirow{2}{*}{16.26} & \multirow{2}{*}{16.95} & \multirow{2}{*}{16.95} & \multirow{2}{*}{16.86} \\

& \#Params       & 0.06  & 0.21  & 0.05  & 3.16 & 3.16  & 0.33 & 3.16  \\ 
\midrule

\multirow{5}{*}{HC} 
& Training Time & \multirow{2}{*}{1839.1} & \multirow{2}{*}{4311.1} & \multirow{2}{*}{2590.1} 
& \multirow{2}{*}{962.1} & \multirow{2}{*}{2005.4} & \multirow{2}{*}{1083.1} & \multirow{2}{*}{1838.1} \\
& Testing Time  & \multirow{2}{*}{5.67}  & \multirow{2}{*}{5.78}  & \multirow{2}{*}{6.48} 
& \multirow{2}{*}{31.03} & \multirow{2}{*}{30.81} & \multirow{2}{*}{30.81} & \multirow{2}{*}{30.82} \\

& \#Params       & 0.08  & 0.23  & 0.07  & 3.42 & 3.42  & 0.34 & 3.42  \\ 
\midrule

\multirow{5}{*}{SA} 
& Training Time & \multirow{2}{*}{1720.5} & \multirow{2}{*}{3984.1} & \multirow{2}{*}{2474.9} 
& \multirow{2}{*}{966.4} & \multirow{2}{*}{2934.3} & \multirow{2}{*}{1064.1} & \multirow{2}{*}{2791.3} \\
& Testing Time  & \multirow{2}{*}{7.15}  & \multirow{2}{*}{6.35}  & \multirow{2}{*}{7.90} 
& \multirow{2}{*}{33.89} & \multirow{2}{*}{34.64} & \multirow{2}{*}{34.64} & \multirow{2}{*}{34.95} \\

& \#Params       & 0.07  & 0.22  & 0.06  & 3.31 & 3.31  & 0.35 & 3.31  \\ 
\midrule

\multirow{5}{*}{IP} 
& Training Time & \multirow{2}{*}{1670.0} & \multirow{2}{*}{3906.7} & \multirow{2}{*}{2417.2} 
& \multirow{2}{*}{673.2} & \multirow{2}{*}{1235.9} & \multirow{2}{*}{996.5} & \multirow{2}{*}{1138.9} \\
& Testing Time  & \multirow{2}{*}{1.32}  & \multirow{2}{*}{1.16}  & \multirow{2}{*}{1.51} 
& \multirow{2}{*}{6.35} & \multirow{2}{*}{6.41} & \multirow{2}{*}{6.41} & \multirow{2}{*}{6.49} \\

& \#Params       & 0.07  & 0.22  & 0.06  & 3.3 & 3.3  & 0.35 & 3.3  \\ 
\bottomrule

\end{tabularx}
\end{table*}

\subsection{Computational Complexity Analysis}
This study compares the computational efficiency of the proposed S4L-FSC method with several representative few-shot learning methods. These compared methods are DCFSL, GIA-CFSL, ADAFSL, HFSL, and FSCF-SSL. We assessed training time, test time, and the number of model parameters. Results are summarized in Table~\ref{tab:complexity}.

Regarding the number of model parameters, S4L-FSC, FSCF-SSL, and HFSL all utilize a common VGG-based spectral-spatial feature extraction network. Consequently, they share nearly identical parameter counts. This demonstrates that the improvements of the S4L-FSC method primarily stem from its training strategy. In contrast, other methods like DCFSL, GIA-CFSL, and ADAFSL employ different network architectures and thus have relatively fewer parameters, with values ranging from $0.05\,\mathrm{M}$ to $0.23\,\mathrm{M}$.

In terms of test time, the shared model architecture of S4L-FSC, FSCF-SSL, and HFSL results in very similar test times across all datasets, as shown in Table~\ref{tab:complexity}. This indicates that S4L-FSC improves classification performance without imposing an additional computational burden during the testing phase. Conversely, methods with fewer parameters, such as DCFSL, GIA-CFSL, and ADAFSL, exhibit faster test speeds but often achieve lower classification accuracy.

Training time reflects the computational overhead associated with the model training process. Methods involving complex pre-training or meta-learning processes, including S4L-FSC and FSCF-SSL, typically require longer training periods than other approaches. When comparing S4L-FSC with FSCF-SSL, S4L-FSC trains faster on the HC, SA, and IP datasets and is only slightly slower on the UP dataset. S4L-FSC incorporates an additional homogeneous data pre-training stage and utilizes data augmentation in the target domain contrastive learning. These factors contribute to a marginally longer training time on the UP dataset compared to FSCF-SSL. However, S4L-FSC's training is faster on the other three datasets. This efficiency gain is due to different batch size settings in the target domain self-supervised contrastive learning stage. The batch size for S4L-FSC in this stage is $N_t \times K_0$. For example, with $K_0=5$, the batch size is $45$ for the UP dataset ($9 \times 5$) and $80$ for the other three datasets ($16 \times 5$). This is smaller than the fixed batch size of 128 used by FSCF-SSL's self-supervised learning component \cite{fscf-ssl}. Using a smaller batch size in this specific stage accelerates iteration updates, partially offsetting the time overhead introduced by other stages. Consequently, the total training time for our method is reduced on certain datasets. 

In summary, the S4L-FSC method achieves improved classification performance through its multi-stage, multi-source, multi-task training approach. This enhancement is realized without altering the model's parameter count or theoretical test time. Although the training process is more complex, the total training time is competitive with the FSCF-SSL method due to optimized batch sizes in specific stages.

\section{CONCLUSION}
In this study, we proposed a novel few-shot learning framework S4L-FSC, which aims to address the challenge of scarce labeled samples in hyperspectral image classification. By fusing SSL and FSL, the method significantly improves the classification performance under limited labeled data. S4L-FSC adopts the pre-training strategy of heterogeneous datasets and homogeneous datasets, and enhances the model's adaptability to the spatial geometric diversity and spectral prior information of hyperspectral images through RM-SSL and MR-SSL, both combined with FSL. In the target domain, we further combine FSL and contrastive learning for fine-tuning to optimize the classification performance.

Experimental results show that S4L-FSC outperforms existing advanced methods on four public hyperspectral image datasets, verifying its superiority and robustness in few-shot scenarios.

\bibliographystyle{IEEEtran}
\bibliography{references} 

\end{document}